\documentclass{article}
\usepackage{graphicx}
\usepackage{caption}
\usepackage{subcaption}
\usepackage{booktabs}
\usepackage{multirow}    
\usepackage{pdfpages}
\usepackage[colorlinks=true, linkcolor=black, citecolor=black, urlcolor=blue]{hyperref}

\usepackage{mdframed}
\usepackage{todonotes}
\usepackage{wrapfig}

\definecolor{eqbg}{gray}{0.95}

\newmdenv[
  backgroundcolor=eqbg,
  linewidth=0pt,
  innerleftmargin=1pt,
  innerrightmargin=1pt,
  innertopmargin=1pt,
  innerbottommargin=1pt,
  skipabove=1pt,
  skipbelow=1pt,
  nobreak=true,            
  font=\small              
]{eqbox}

\AtBeginEnvironment{eqbox}{%
  \setlength{\abovedisplayskip}{2pt}%
  \setlength{\belowdisplayskip}{2pt}%
  \setlength{\abovedisplayshortskip}{2pt}%
  \setlength{\belowdisplayshortskip}{2pt}%
}

\usepackage[final]{neurips_2025}

\usepackage{makecell}
\usepackage[utf8]{inputenc} 
\usepackage[T1]{fontenc}    
\usepackage{hyperref}       
\usepackage{url}            
\usepackage{booktabs}       
\usepackage{amsfonts}       
\usepackage{nicefrac}       
\usepackage{microtype}      
\usepackage{xcolor}         
\usepackage{graphicx}
\usepackage{algorithm}%
\usepackage{algorithmic}
\usepackage{amsmath}
\usepackage{cleveref}
\usepackage{multirow}
\usepackage{graphicx}
\usepackage{enumitem}
\newcommand{\ie}{i.e., }

\newcommand{\bfA}{{\bf A}}

\newcommand{\bfI}{{\bf I}}
\newcommand{\bfJ}{{\bf J}}

\newcommand{\bfL}{{\bf L}}

\newcommand{\bfW}{{\bf W}}
\newcommand{\bfX}{{\bf X}}

\newcommand{\bfx}{{\bf x}}

\usepackage{amsthm}
\newtheorem{theorem}{Theorem}
\newtheorem{lemma}{Lemma}[section]

\title{Return of ChebNet: Understanding and Improving an Overlooked GNN on Long-Range Tasks}

\author{%
  \makebox[\textwidth][c]{%
    \textbf{Ali Hariri}\textsuperscript{1,*,\dag} \quad
    \textbf{Álvaro Arroyo}\textsuperscript{2,*} \quad
    \textbf{Alessio Gravina}\textsuperscript{3,*} \quad
    \textbf{Moshe Eliasof}\textsuperscript{4}%
  }\\[1ex]
  \makebox[\textwidth][c]{%
    \textbf{Carola‐Bibiane Schönlieb}\textsuperscript{4} \quad
    \textbf{Davide Bacciu\textsuperscript{3}} \quad
    \textbf{Kamyar Azizzadenesheli\textsuperscript{5}} 
  }\\[1ex]
  \makebox[\textwidth][c]{%
    \textbf{Xiaowen Dong\textsuperscript{2}} \quad
    \textbf{Pierre Vandergheynst\textsuperscript{1}}%
  }
}

\begin{document}

\maketitle

\renewcommand\thefootnote{\arabic{footnote}}  
\setcounter{footnote}{0}  
\footnotetext[1]{École Polytechnique Fédérale de Lausanne (EPFL)}
\footnotetext[2]{University of Oxford}
\footnotetext[3]{University of Pisa}
\footnotetext[4]{University of Cambridge}
\footnotetext[5]{NVIDIA Research}

\renewcommand\thefootnote{\fnsymbol{footnote}}
\setcounter{footnote}{0}
\footnotetext[1]{Equal contribution}
\footnotetext[2]{Correspondence to: Ali Hariri (\texttt{ali.hariri@epfl.ch})}
\renewcommand\thefootnote{\arabic{footnote}}  

\begin{abstract}
ChebNet, one of the earliest spectral GNNs, has largely been overshadowed by Message Passing Neural Networks (MPNNs), which gained popularity for their simplicity and effectiveness in capturing local graph structure. Despite their success, 
MPNNs are 
limited in their ability to capture long-range dependencies 
between nodes. This has led researchers to adapt MPNNs through \textit{rewiring} or make use of \textit{Graph Transformers}
, which compromises the computational efficiency that characterized early spatial message-passing architectures, and typically disregards the graph structure. Almost a decade after its original introduction, we revisit ChebNet to shed light on its ability to model distant node interactions. We find that out-of-box, ChebNet already shows competitive advantages relative to classical MPNNs and GTs on long-range benchmarks, while maintaining good scalability properties for high-order polynomials. However, we uncover that this polynomial expansion leads ChebNet to an unstable regime during training. To address this limitation, we cast ChebNet as a stable and non-dissipative dynamical system
, which we coin \texttt{Stable-ChebNet}. Our \texttt{Stable-ChebNet} model allows for stable information propagation, and has controllable dynamics which do not require the use of eigendecompositions, positional encodings, or graph rewiring. Across 
several benchmarks, \texttt{Stable-ChebNet} achieves near  state-of-the-art performance.


\end{abstract}

\section{Introduction}

\vspace{-5pt}

Graph Neural Networks (GNNs) \citep{sperduti1993encoding, gori2005new, scarselli2008graph, NN4G, bruna2013spectral, defferrard2016convolutional,gravina_dynamic_survey} have emerged as a prevalent framework for handling data defined on graphs. Graph convolutional networks have their roots in spectral approaches that extend convolutional filters to non-Euclidean domains. The first practical instantiation of a GNN was proposed by \citep{bruna2013spectral}, which leveraged the eigenbasis of the graph Laplacian to perform spectral filtering, as an attempt to generalize image convolutions to non-euclidean structures. However, this formulation required costly eigen-decompositions at each layer. Defferrard \emph{et al.}~\cite{defferrard2016convolutional} addressed this inefficiency by approximating spectral filters with truncated Chebyshev polynomials, giving rise to \emph{ChebNet}, the first tractable and localized spectral GNN. By parameterizing filters as \(K\)-order polynomials of the Laplacian, ChebNet could aggregate information from \(K\)-hop neighborhoods without repeated eigendecompositions, thereby enabling scalable spectral convolution on large graphs.

In 2017, Kipf and Welling distilled ChebNet into a simpler, first-order approximation now known as the \emph{Graph Convolutional Network (GCN)} by (i) restricting the polynomial order to one and (ii) tying filter coefficients across hops~\cite{kipf2016semi}. This yielded an architecture that was both lightweight and effective: a GCN with few layers would achieve strong node‐classification performance on standard homophilic benchmarks, and its \(\mathcal{O}(\lvert E\rvert)\) complexity made it practical for large-scale graphs. GCN’s efficiency and strong locality bias quickly made it the default baseline, and subsequent \emph{message-passing neural networks (MPNNs)} adopted a similar paradigm of iterative neighborhood aggregation~\cite{gilmer2017neural}.

\begin{wrapfigure}{r}{0.65\linewidth}
    \centering
    \vspace{-10pt}
    \includegraphics[width=\linewidth]{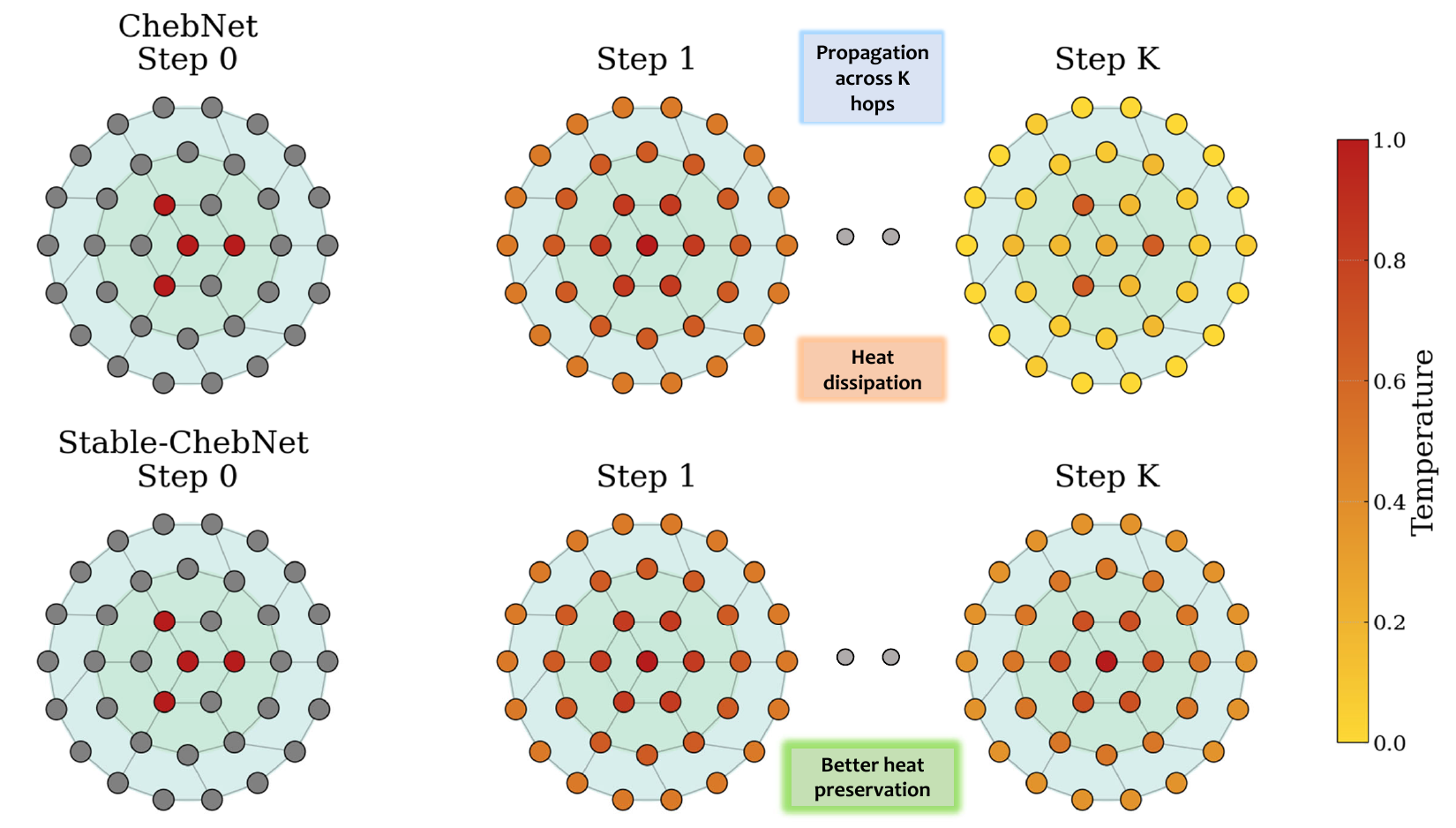}
    \caption{\textbf{Top:} Vanilla ChebNet. \textbf{Bottom:} Stable-ChebNet. While the original ChebNet’s high-order Chebyshev filters induce unstable dynamics resulting in dissipative behavior, our Stable-ChebNet yields bounded propagation through layers. }
    \vspace{-10pt}
    \label{fig:main_figure}
\end{wrapfigure}

Despite their popularity, MPNNs exhibit pronounced shortcomings when made deeper and when capturing long-range dependencies \cite{dwivedi2022long}. Repeated neighborhood aggregations tend to cause \emph{representational collapse}, where node features become indistinguishable (often referred to as “oversmoothing”)~\cite{cai2020note,oono2020graph}, and information from distant nodes is “squashed” through narrow bottlenecks, limiting the ability to model global context~\cite{alon2021on}. In recent years, researchers have proposed various strategies to overcome these limitations. To mitigate oversmoothing, several models have drawn on principles from physics \cite{di2022graph,bodnar2022neural} to preserve feature diversity across layers. At the same time, efforts to capture long-range dependencies have led to graph-rewiring techniques \cite{topping2022understanding,gutteridge2023drew,barbero2023locality} that add or reweight edges to shorten information pathways, as well as the emergence of graph transformers \cite{dwivedi2020generalization}, which replace purely local aggregation with global self-attention mechanisms. Although these advances can alleviate depth-related pathologies, they often trade off numerous benefits that made MPNNs appealing: scalability, parameter efficiency, and to only process information along the graph’s edges. In this context, ChebNet and other spectral GNNs are usually relegated to a footnote - mentioned only as a predecessor to GCN, which is typically not revisited as a competitive baseline.

In this work, we revisit ChebNet from first principles. We demonstrate that the original ChebNet (without any rewiring or attention mechanisms) already delivers state-of-the-art performance or is close on long-range graph tasks while scaling gracefully to large graphs. By deriving and analyzing ChebNet’s linearized dynamics, we prove that enlarging its receptive field introduces signal-propagation instabilities. To overcome this, we propose \texttt{Stable-ChebNet}, a minimal set of architectural modifications that restore stable propagation for arbitrarily large receptive fields, supported by both theoretical guarantees and empirical validation. To build intuition for our Stable-ChebNet framework, Figure \ref{fig:main_figure} illustrates how classical ChebNet filters (top) can exhibit unbounded dynamics, whereas our antisymmetric, forward-Euler discretization yields smooth, stable propagation (bottom). An intuitive heat-transfer analogy helps explain the difference: if we inject ‘heat’ at seed nodes, the high-order filters of vanilla ChebNet diffuse and dissipate this heat so that distant nodes cool rapidly. In contrast, the non-dissipative dynamics induced by the antisymmetric, forward-Euler step in Stable-ChebNet preserve energy, keeping temperatures higher at nodes many hops away. 

Across a suite of challenging long-range node- and graph-level benchmarks, \texttt{Stable-ChebNet} matches or outperforms state-of-the-art message-passing neural networks and graph transformers, while retaining the ChebNet backbone. We hope this work will reignite interest in spectral GNNs as a scalable, theoretically grounded alternative for long-range graph modeling.
\vspace{-5pt}
\paragraph{Contributions and Outline}
\vspace{-5pt}

\begin{itemize}[left=5pt, topsep=0pt, partopsep=0pt, itemsep=2pt, parsep=0pt]
    \item In Section \ref{sec:empirical-effectiveness}, we empirically demonstrate that vanilla ChebNet can achieve very strong performance on long-range benchmarks without incurring prohibitive computational cost. 
    \item In Section \ref{sec:signal-propagation}, we analyze ChebNet’s signal‐propagation dynamics, providing exact sensitivity analysis, and theoretically and empirically prove the emergence of instability for large filter order.
    \item In Section \ref{sec:stable-cheb}, we introduce \texttt{Stable-ChebNet}, which enforces layer-wise stability.
    \item  In Section \ref{sec:experiments}, we empirically validate that \texttt{Stable-ChebNet} consistently outperforms MPNNs, rewiring methods, and graph transformers across a number of tasks.
\end{itemize}

\vspace{-10pt}

\section{Background} \label{background}

\vspace{-5pt}

\subsection{Background on Spectral Graph Neural Networks}

\vspace{-5pt}

Spectral 
GNNs extend the notion of convolution to graphs by leveraging the eigen-decomposition of the graph Laplacian. Given an undirected graph $G = (V, E)$ with normalized Laplacian $\textbf{L} = \mathbf{I} - \mathbf{D}^{-1/2} \mathbf{A} \mathbf{D}^{-1/2}$, any graph signal $\bfX \in \mathbb{R}^n$ can be filtered in the spectral domain via \( \mathbf{Y} = \mathbf{U} g_{\boldsymbol{\theta}}(\boldsymbol{\Lambda}) \mathbf{U}^\top \mathbf{X} \), where \( \mathbf{L} = \mathbf{U} \boldsymbol{\Lambda} \mathbf{U}^\top \) diagonalizes the Laplacian, 
$\boldsymbol{\Lambda} = \text{diag}(\lambda_1, \dots, \lambda_n)$ its eigenvalues, and $g_{\boldsymbol{\theta}}$ is a learnable spectral response. Early methods directly parameterize $g_{\boldsymbol{\theta}}(\Lambda)$, but require an expensive eigendecomposition of the Laplacian, which can be computationally and memory intensive \cite{bruna2013spectral}. ChebNet alleviates the cost of an explicit eigendecomposition by approximating $g_{\boldsymbol{\theta}}$ using the recurrence relation for a $K$-th order Chebyshev polynomial in $\bfL$ \cite{defferrard2016convolutional}.  The latter defines $g_{\boldsymbol{\theta}}$ as $g_{\boldsymbol{\theta}}(\Lambda)\approx\sum_{k=0}^K \boldsymbol{\Theta}_k\,T_k(\tilde\Lambda)$, where $T_k(\tilde\Lambda)$ is the $k$-th polynomial of $\tilde\Lambda$ with $\tilde\Lambda =  \frac{2\Lambda}{\lambda_{max}}-\bfI_n$ .  
The spectral convolution can then be written without any eigendecomposition as the truncated expansion:
\vspace{-5pt}
\begin{equation}\label{eq:cheb-conv}
\mathbf{Y} \;=\;\sum_{k=0}^K\boldsymbol{\Theta}_k\,T_k(\tilde {\mathbf{L}})\,\mathbf{X} 
\end{equation}
\vspace{-5pt}
where $\tilde{\bfL} =  \frac{2\bfL}{\lambda_{max}}-\bfI_n$
enabling efficient, localized filtering in $\mathcal O(K|E|)$ time. 


\subsection{MPNNs and their Limitations} 
Message-Passing Neural Networks (MPNNs) define a general framework in which node features are iteratively updated by exchanging ``messages'' along edges. At each layer $l$, every node $v$ aggregates information from its neighbors $u \in \mathcal{N}(v)$ and combines it with its own representation. 
This formulation unifies many graph models, including graph convolutional networks (GCNs) \cite{kipf2016semi} and graph attention networks (GATs) \cite{velivckovic2018graph}. While this local neighborhood aggregation captures structural information effectively, it has a limited capacity to model long-range interactions within the graph. This is due to the phenomenon of \textit{over-squashing}, an information bottleneck that impedes effective information flow among distant nodes \cite{alon2021on,topping2022understanding,di2023over}. Numerous techniques have emerged to address this limitation such as graph rewiring \cite{gutteridge2023drew, barbero2023locality}, Graph Transformers \cite{rampavsek2022recipe,shirzad2023exphormer,shirzad2024even} in addition to some enhanced spatial methods that tackle over-squashing through combined local and global information \cite{sestak2024vn,geisler2024spatio}, or through non-dissipativity achieved by antisymmetric weight parameterization \cite{gravina_adgn, gravina_swan} or port-Hamiltonian systems \cite{gravina_phdgn}. 

\begin{wrapfigure}{r}{0.28\linewidth}
  \centering
  \vspace{-25pt}
  \includegraphics[width=\linewidth]{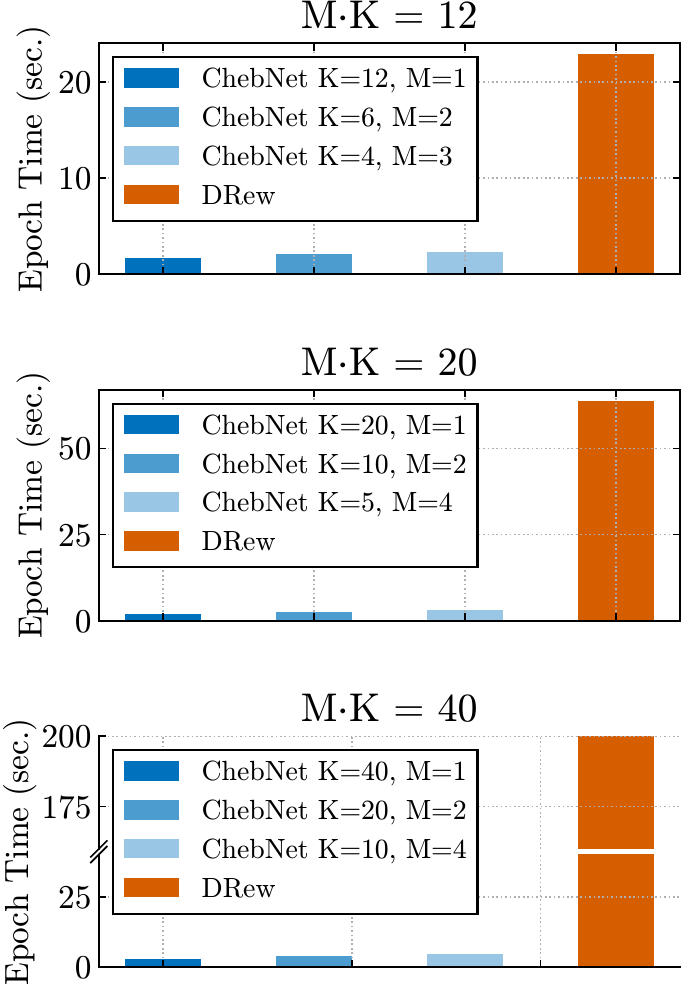}
  \caption{Epoch times for DRew and ChebNet for different receptive fields on the peptides-func task. M is the number of layers, and K is the number of filters.}
  \vspace{-15pt}
  \label{fig:runtime-comparison}
\end{wrapfigure}

However, some of the aforementioned methods suffer from substantial overhead due to denser graph shift operators or the use of all-pairs interactions. Specifically, \cite{shirzad2024even} and other graph transformers increase computational complexity through dense attention-maps; \cite{geisler2024spatio} relies on costly full eigendecomposition operations; and graph rewiring techniques heavily pre-process the graph topology, incurring $\mathcal{O}(n^3)$ time in the case of \cite{gutteridge2023drew}.For a detailed discussion on the relevant literature, we point the reader to the Appendix~\ref{app:supplementary_related_work}.

\vspace{-10pt}
\subsection{The Connection between ChebNet and GCN}

\vspace{-10pt}

Although often interpreted as disparate models, GCN \cite{kipf2016semi} can be derived as a special case of ChebNet \cite{defferrard2016convolutional} by truncating the Chebyshev expansion to $K=1$ and making additional simplifications such as approximating $\tilde{\mathbf{L}} \approx \mathbf{I} - \mathbf{D}^{-1/2} \mathbf{A} \mathbf{D}^{-1/2}$ and adding self-loops to improve numerical stability. These choices introduced a strong locality bias, which aligned with widely used homophilic benchmarks and significantly reduced computational costs. Due to some of these strengths, GCN has become the de facto backbone for many modern GNNs, and has led to ChebNet being somewhat forgotten, under the assumption that it does not perform well and will not scale well in mid- to large-size graphs due to its spectral nature. For instance, until recently, ChebNet was almost never included in popular GNN benchmarks, such as \cite{gnnbenchmark}, or in those designed to evaluate long-range dependencies, such as those in the Long Range Graph Benchmark (LRGB) \cite{dwivedi2022long} and the numerous studies that leveraged this dataset to assess the effectiveness of graph rewiring, positional encodings, or Transformer-based architectures. ChebNet was similarly disregarded from large-scale graph evaluations such as the OGB benchmarks \cite{ogb}, presumably due to prevailing assumptions about its scalability.

\vspace{-5pt}
\section{Analyzing and Improving ChebNet from First Principles}
\vspace{-5pt}

\subsection{The Effectiveness and Scalability of Vanilla ChebNet}\label{sec:empirical-effectiveness}
\vspace{-5pt}

\begin{wrapfigure}{l}{0.25\linewidth}
  \centering
  \vspace{-6pt}
  \includegraphics[width=\linewidth]{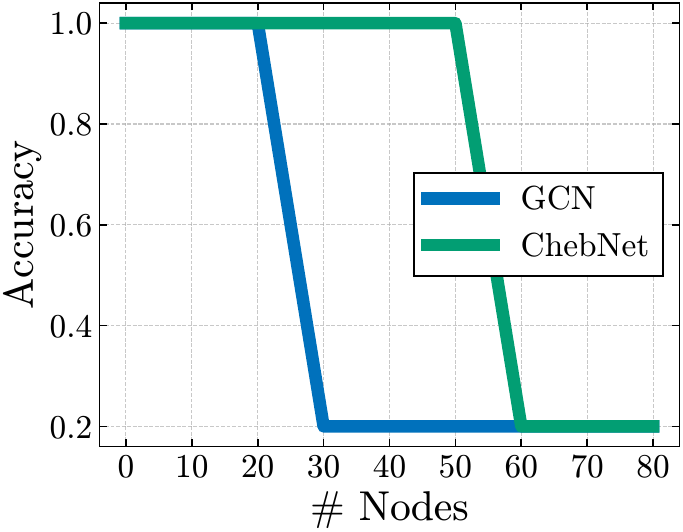}
  \caption{Test accuracy on RingTransfer.}
  \vspace{-15pt}
  \label{fig:ringtransfer-comparison}
\end{wrapfigure}

In this subsection, we perform two high-level empirical tests to 
challenge the commonplace assumption that spectral GNNs inherently suffer from poor performance and limited scalability. We do so by firstly testing ChebNet \cite{defferrard2016convolutional} on a long-range test on the Ring Transfer dataset from \cite{di2023over}, which has become a de facto benchmark for state-of-the-art methods seeking to model long-distance dependencies on graphs. Furthermore, to test scalability, we compare epoch training times on the \texttt{peptides-func} dataset from \cite{dwivedi2022long} with respect to state-of-the-art rewired MPNNs \cite{gutteridge2023drew} based on a GCN backbone. We compare ChebNet with different numbers of filters $K$ and layers $M_{\text{Cheb}}$ with an $M_{\text{MPNN}}$-layer MPNN. To ensure a fair comparison, we ensure that $M_{\text{Cheb}}K = M_{\text{MPNN}}$ so that the two methods would have the same receptive field. The results are shown in Figures \ref{fig:runtime-comparison} and  \ref{fig:ringtransfer-comparison}.

As seen from \Cref{fig:ringtransfer-comparison}, ChebNet is capable of performing long-range retrieval on the ring transfer dataset for rings of up to 50 nodes, which is a substantial improvement over a regular GCN. While this finding aligns with intuition, as we will formalize in the following sections, it may 
nonetheless surprise practitioners, since ChebNet is not typically included as a benchmark in long-range studies such as that in \cite{dwivedi2020generalization}. On the other hand, as seen in Figure \ref{fig:runtime-comparison}, ChebNet scales gracefully on standard tasks such as the \texttt{peptides-func} dataset.  When compared to DRew, a state-of-the-art method that leverages both dynamic rewiring and delay to propagate information through a GCN (or other MPNN) backbone, we observe that DRew’s epoch times are almost up to two orders of magnitude larger than ChebNet’s.  We note further that this inefficiency is not unique to this particular baseline: Graph Transformers incur quadratic scaling in the number of nodes and, by effectively discarding the underlying graph structure, trade-off inductive bias for increased compute, while many rewiring approaches depend on cubic-time algorithms (e.g., Floyd–Warshall or eigendecompositions). Together, these examples underscore how numerous contemporary techniques intended to overcome traditional message-passing limitations actually erode computational benefits, whereas ChebNet, a more natural spectral baseline that generalizes GCN, delivers both training speed and strong performance out-of-the-box.

\vspace{-10pt}

\subsection{Signal Propagation Analysis of ChebNet}\label{sec:signal-propagation}

\vspace{-5pt}
In this subsection, we conduct a sensitivity analysis of ChebNet via the spectral norm of the Jacobian of node features, providing an exact characterization of ChebNet's information flow through different layers and between pairs of nodes, in the spirit of \cite{arroyo2025vanishing}. Specifically, we begin by analyzing the layer-wise Jacobian for Spectral GNNs that use polynomial filters. In this setting, we demonstrate that the layer-wise Jacobian becomes unstable as the polynomial order $K$ increases. Lastly, we investigate the sensitivity of node pairs when using ChebNet. We provide the proofs for the statements in \Cref{app:proofs}. \\


\begin{eqbox}
\begin{lemma}[Layer-Wise Jacobian for a Spectral GNN]
	\label{lem:linear_jacobian}
    Consider 
    a linear spectral GNN whose layer-wise update is performed through the following polynomial filter 
    $f(\mathbf{X}) = \sum_{k=1}^{K} T_k(\bfL) \, \mathbf{X} \, \Theta_k$, where \( \mathbf{X} \in \mathbb{R}^{n \times d} \) is the node feature matrix, 
    $T_k(\bfL)\in\mathbb{R}^{n\times n}$ is the $k$-th polynomial of the Laplacian $\bfL\in\mathbb{R}^{n\times n}$, and \( \mathbf{\Theta}_k \in \mathbb{R}^{d \times d'} \) are learnable weight matrices. Then, the vectorized Jacobian $\mathbf{J} = \partial\operatorname{vec}(f(\mathbf{X}))/\partial\operatorname{vec}(\mathbf{X})$ is 
    \begin{equation}\label{eq:spectral_gnn_jacobian}
    \mathbf{J} = \sum_{k=1}^{K} \mathbf{\Theta}_k^\top \otimes T_k(\bfL) .
    \end{equation}
\end{lemma}
\end{eqbox}
\looseness=-1

Given the layer-wise Jacobian from Lemma \ref{lem:linear_jacobian}, we proceed to analyze the dynamics of a Spectral GNN in Theorem \ref{lem:nonlinear_jacobian} below. Specifically, we focus on the case where the polynomial filter can be approximated by powers of the Laplacian, \ie $T_k(\bfL)=\bfL^k$.\\ 

\begin{eqbox}
\begin{theorem}[Layer-Wise Jacobian singular-value distribution]
	\label{lem:nonlinear_jacobian}
    Assume the setting of Lemma~\ref{lem:linear_jacobian}, with $T_k(\bfL)=\bfL^k$ and $\bfL$ the symmetric normalized Laplacian, and let all
$\boldsymbol{\Theta}_k\in\mathbb{R}^{d\times d}$ be initialized with i.i.d.\ 
$\mathcal{N}(0,\sigma^2)$ entries. Denote the  eigenvalues of $\mathbf{L}$ as $\{\lambda_1,\ldots,\lambda_n\}$, the squared singular values of $\mathbf{\Theta}_k\,\mathbf{\Theta}_k^T$ as $\{\mu_{1,k},\ldots,\mu_{d,k}\}$, and the squared singular values of the 
Jacobian by $\gamma_{i,j}$. Then, for sufficiently large $d$ the empirical eigenvalue distribution of $\boldsymbol{\Theta}_k\boldsymbol{\Theta}_k^T$ converges to the 
Marchenko-Pastur distribution $\forall k$. Then, 
the mean and variance of each $\gamma_{i,j}$ are
\begin{align}
  \mathbb{E}\bigl[\gamma_{i,j}\bigr]
  &= 
  \sigma^2\,\sum_{k=1}^K \lambda_i^{2k} ,
  \label{eq:exp} \\[2pt]
  \mathrm{Var}\bigl[\gamma_{i,j}\bigr]
  &=
  \sigma^4\,\bigg(\sum_{k=1}^K \lambda_i^{2k}\bigg)^2.
  \label{eq:var}
\end{align}
\end{theorem}
\end{eqbox}

\begin{figure}[t]
    \centering
    \includegraphics[width=1\linewidth]{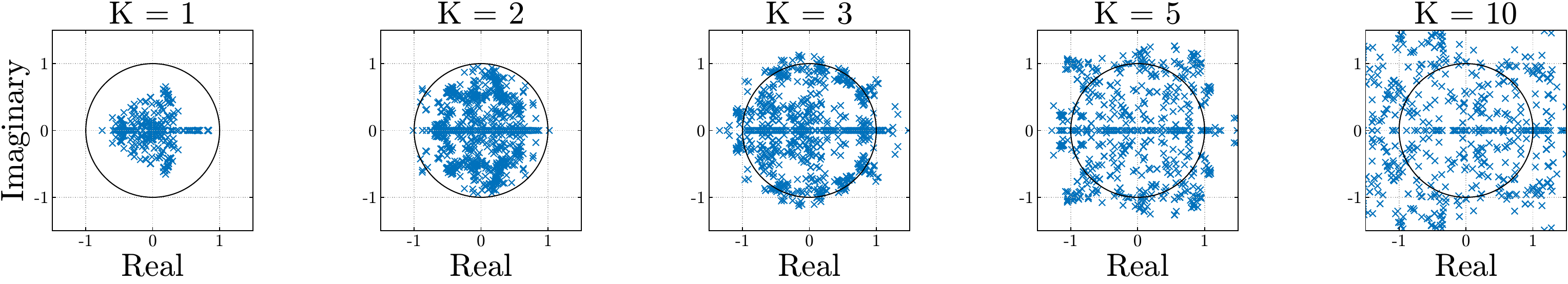}
    \caption{Singular‐value spectra of the graph‐wise Jacobian in the complex plane for vanilla ChebNet with increasing polynomial order \(K\).}
    \vspace{-12pt}
    \label{fig:jacobian-k}
\end{figure}

\Cref{lem:nonlinear_jacobian} shows that the singular values spectrum of the layer-wise Jacobian depends on the sum over the powers of the normalized Laplacian's eigenvalues. This indicates that larger polynomial orders $K$ push the singular value-spectrum towards unstable dynamics, as empirically demonstrated in \Cref{fig:jacobian-k}. Stacking several layers of large filter orders will therefore severely hinder the trainability of a Spectral GNN of this form.

Beyond analyzing the information propagation dynamics between different layers, we are interested in quantifying the communication ability between distant nodes in the graph. Recent literature has proposed to measure information flow in the graph by evaluating the sensitivity of a node embedding after $l$ layers (\ie hops of propagation) with respect to the input of another node using the node-wise Jacobian \cite{topping2022understanding,di2023over, arroyo2025vanishing}, \ie $\partial\bfx_u^{(l)} / \partial\bfx_v^{(0)}$. Following this approach, we measure how sensitive  
a node embedding of ChebNet at an arbitrary layer $l$ with respect to the initial features of another node, to better illustrate the long-range propagation capabilities of spectral GNNs. 
\vspace{0.2cm}
\begin{eqbox}    
\begin{theorem}[ChebNet Sensitivity]\label{thm:chebnet_sensitivity}
Consider a Chebyshev-based Graph Neural Network (ChebNet) defined as:
\begin{equation}
    \mathbf{X}^{(l+1)} = \sum_{k=0}^{K} T_k(\mathbf{L}) \mathbf{X}^{(l)} \mathbf{W}_k^{(l)},
\end{equation}
where $\mathbf{L}\in \mathbb{R}^{n\times n}$ is the graph Laplacian, $T_k(\mathbf{L})$ is the $k$-th Chebyshev polynomial of the Laplacian, and $\mathbf{W}_k^{(l)}$ are learnable weight matrices. Assume activation function $\sigma$ is identity and let $\mathbf{X}^{(0)}$ be the input features. 

Then, the sensitivity of node $v$ with respect to node $u$ after $l$ layers is given by:
\begin{equation}
    \frac{\partial\mathbf{x}_v^{(l)}}{\partial\mathbf{x}_u^{(0)}} = \left(\prod_{l=0}^{l-1} \left(\sum_{k=0}^{K} T_k(\mathbf{L})\mathbf{W}_k^{(l)}\right)\right)_{v,u}.
\end{equation}
\end{theorem}
\end{eqbox}

This result indicates that the sensitivity of ChebNet is closely tied to the polynomial order $K$. We note that, in the case of $K=1$, the sensitivity aligns with that of standard MPNNs \footnote{The sensitivity upper bound of standard MPNNs is further discussed in \Cref{app:sensitivity_mpnns}.} (such as GCN), exhibiting reduced long-range communication. However, for $K>1$, the higher polynomial orders significantly enhance the ChebNet's sensitivity, enabling more effective long-range propagation and improving the overall capacity to capture distant dependencies in the graph. Therefore, we conclude that a large filter order K is needed to enable long-range communication between nodes in the graph, but this will result in unstable training dynamics. This serves to explain the decay in performance in Figure \ref{fig:ringtransfer-comparison}. In the next subsection, we will propose a remedy to this issue. 

\vspace{-7pt}

\subsection{Stable-ChebNet:  Stability with Antisymmetric Parameterization} \label{sec:stable-cheb}
\vspace{-3pt}

As discussed in \Cref{sec:signal-propagation}, although ChebNet demonstrates strong long-range propagation capabilities, increasing the polynomial order can introduce significant instability into the model dynamics. 
To address this challenge, we propose Stable-ChebNet, a simple yet effective modification of classical ChebNet aimed at improving its stability. Recent literature has demonstrated that the effectiveness of neural architectures can be significantly improved by framing them as stable, non-dissipative dynamical systems \cite{HaberRuthotto2017, chang2018antisymmetricrnn, gravina_adgn, gravina_swan, gravina_phdgn}. The core idea behind these approaches is to carefully regulate the spectrum of the Jacobian matrix to ensure the network operates within a stable regime. Specifically, this behavior can be achieved by constraining the eigenvalues of the Jacobian to be purely imaginary. Under this constraint, the input graph information is effectively propagated through the successive transformations into the final nodes’ representation. Motivated by this line of work, we begin by reformulating ChebNet as a continuous-time differential equation. Specifically, we consider the following ordinary differential equation (ODE):
\begin{equation}\label{eq:chebnetODE}
    \frac{d\bfX(t)}{dt} = \sum_{k=0}^K T_k(\bfL)\bfX(t)\bfW_k
\end{equation}
for time $t\in[0,T]$ and subject to the initial condition (\ie the input features) $\bfX(0)=\bfX^{(0)}$. In other words, the dynamics of the system (\ie the continuous flow of information over the graph) is now described as the ChebNet update rule.
To ensure the Jacobian of this system has purely imaginary eigenvalues, a straightforward approach is to use antisymmetric weight matrices\footnote{A matrix $\bfA\in\mathbb{R}^{d\times d}$ is antisymmetric (\ie skew-symmetric) if $-\bfA = \bfA^\top$} and the symmetrically normalized Laplacian. This choice, as formalized in the following theorem, directly enforces the desired spectral property, leading to inherently stable dynamics.
\\
\begin{eqbox}
\begin{theorem}[Purely Imaginary Eigenvalues]\label{thm:chebnetODE_eigenvalues}
Let $\bfL$ be the symmetric normalized Laplacian and $-\bfW_k = \bfW_k^\top$ $\forall k=0,\dots,K$, then the graph-wise Jacobian of the ODE in \Cref{eq:chebnetODE} has purely imaginary eigenvalues, \ie
\begin{equation}
    Re(\lambda_i(\bfJ)) = 0, \; \forall i.
\end{equation}
\end{theorem}
\end{eqbox}

 Even in this section, we provide the proofs for the statements in \Cref{app:proofs}. \Cref{thm:chebnetODE_eigenvalues} shows that by enforcing antisymmetry in the weight matrices and leveraging the symmetric structure of the Laplacian, we guarantee that the Jacobian has purely imaginary eigenvalues, ensuring that node representations remain sensitive to input features of far away nodes without suffering from the instability of standard ChebNet.

As for standard differential-equation-inspired neural architectures, a numerical discretization method is needed to solve \Cref{eq:chebnetODE}. We solve the equation with a simple finite difference scheme, \ie forward Euler's method, yielding the following node update equation
\begin{equation}\label{eq:stable_chebnet}
    \bfX^{(l+1)} =\bfX^{(l)} + \epsilon \left(\sum_{k=0}^K T_k(\bfL)\bfX^{(l)}(\bfW_k-\bfW_k^\top-\gamma\bfI)\right)
\end{equation} 
where $\bfI$ is the identity matrix, $\gamma\in\mathbb{R}$ is a hyper-parameter that maintains the stability of the forward Euler method, and $\epsilon\in\mathbb{R}_+$ is the discretization step.

We refer to the ODE in \Cref{eq:stable_chebnet} as Stable-ChebNet, and in the following, we show that this new formulation achieves second-order stability. This is a critical improvement, as a naive Euler discretization of the original ChebNet (without imposing constraints on the Jacobian eigenvalues) would result in only first-order stability, which is considerably more prone to numerical instability. Specifically, without these constraints, the model can exhibit exponential growth or decay in the gradients, significantly limiting its ability to capture long-range dependencies \cite{arroyo2025vanishing}. In contrast, second-order stable systems, like Stable-ChebNet, maintain controlled gradient dynamics over longer timescales, allowing for effective long-range information propagation.
\vspace{0.25cm}
\begin{eqbox}
\begin{theorem}[Non-exponential Information Growth or Decay with Antisymmetric Weights]\label{thm:Non-exponential_Information_Growth}
Consider the Stable-type Chebyshev Graph Neural Network (Stable-ChebNet) defined by:
\begin{equation}
    \mathbf{X}^{(l+1)} = \mathbf{X}^{(l)} + \epsilon \sum_{k=0}^{K} T_k(\mathbf{L}) \mathbf{X}^{(l)} \mathbf{W}_k^{(l)},
\end{equation}
with small step size $\epsilon > 0$ and antisymmetric weight matrices:
\begin{equation}
    (\mathbf{W}_k^{(l)})^\top = -\mathbf{W}_k^{(l)}, \quad \forall k,l.
\end{equation}
Then the Jacobian $J^{(l)}$ of the layer does not lead to exponential growth or decay across layers. Specifically, we have:
\begin{equation}
    \|\bfJ^{(l)}\|_2 = 1 + O(\epsilon^2).
\end{equation}

Conversely, for general weights without the antisymmetric property, exponential growth or decay of the Jacobian norm typically occurs.
\end{theorem}
\end{eqbox}

\section{Experiments} \label{sec:experiments}

We evaluate ChebNet and its stable formulation (\textbf{Stable-ChebNet}) across a variety of settings to thoroughly assess long-range capabilities. In this section, we report and discuss the performance of both models on these benchmarks. We report additional experiments on heterophilic node classification tasks from \cite{platonov2023a} in Appendix~\ref{app:hetero}. We run our experiments on a single A100 GPU and provide the full details on the hyperparameter search for all datasets in Appendix \ref{append:hyperparams}, and baseline and datasets details in \Cref{app:datasets_and_baselines}. 

\textbf{Graph Property Prediction Dataset.} We evaluate our model's ability to predict long-range graph properties using a synthetic dataset developed by Corso et.al in \cite{corso2020principal} under the experimental setup of \cite{gravina_adgn}. The dataset consists of undirected graphs drawn from a diverse set of random and structured families (Erd\H{o}s--R\'enyi, Barab\'asi--Albert, caterpillar, etc), ensuring a broad coverage of topological properties. Each graph contains between 25 and 35 nodes as per the setup of Gravina et.al in \cite{gravina_adgn} (in contrast to the 15--25 node range originally used by \cite{corso2020principal}), thus increasing task complexity and raising the need for long-range information propagation. Finally, each node is assigned a single scalar feature sampled uniformly at random from the interval $[0, 1]$. We provide a detailed comparison in Table \ref{tab:results_graphprop1}.

\textbf{Results.} Compared to classical ChebNet, Stable-ChebNet yields consistent and significant gains across all three tasks. On the \textit{Diameter} task, classical ChebNet achieves a $\log_{10}(\text{MSE})$ of  $-0.15$, whereas Stable-ChebNet improves it to $-0.25$. On the \textit{Single Source Shortest Path (SSSP)} task, the gain is larger and goes from $-1.85$ using ChebNet to $-2.21$ with our Stable-ChebNet form. Finally, on \textit{Eccentricity}, where the difficulty is highest due to the necessity to propagate information about the most distant nodes individually, classical ChebNet achieves a $\log_{10}(\text{MSE})$ of $-1.22$ while Stable-ChebNet reaches $-2.10$, reducing the average prediction error by more than an order of magnitude relative to the baseline. Relative to other baselines, Stable-ChebNet dominates most models based on standard message-passing or modified diffusion mechanisms. Methods like GCN and GCNII are badly over-squashed, barely reaching negative $\log_{10}(\text{MSE})$ values.
\begin{table}[t]
    \centering
    \caption{Mean test set $\log_{10}(\text{MSE})$ and standard deviation averaged over 4 random weight initializations for each configuration on the Graph Property Prediction dataset. The lower the better.}
    \label{tab:results_graphprop1}
    \begin{tabular}{lccc}
        \toprule
        Model & Diameter & SSSP & Eccentricity \\
        \midrule
        GCN & $0.7424 \pm 0.0466$ & $0.9499 \pm 0.0001$ & $0.8468 \pm 0.0028$ \\
        GAT & $0.8221 \pm 0.0752$ & $0.6951 \pm 0.1499$ & $0.7909 \pm 0.0222$ \\
        GraphSAGE & $0.8645 \pm 0.0401$ & $0.2863 \pm 0.1843$ & $0.7863 \pm 0.0207$ \\
        GIN & $0.6131 \pm 0.0990$ & $-0.5408 \pm 0.4193$ & $0.9504 \pm 0.0007$ \\
        GCNII & $0.5287 \pm 0.0570$ & $-1.1329 \pm 0.0135$ & $0.7640 \pm 0.0355$ \\
        DGC & $0.6028 \pm 0.0050$ & $-0.1483 \pm 0.0231$ & $0.8261 \pm 0.0032$ \\
        GRAND & $0.6715 \pm 0.0490$ & $-0.0942 \pm 0.3897$ & $0.6602 \pm 0.1393$ \\
        A-DGN \scalebox{.7}{w/ GCN backbone}  & $0.2271 \pm 0.0804$ & $-1.8288 \pm 0.0607$ & $0.7177 \pm 0.0345$ \\
        \midrule
        ChebNet & -0.1517 $\pm$ 0.0343 &  -1.8519 $\pm$ 0.0539 &  -1.2151 $\pm$ 0.0852 \\
        \textbf{Stable-ChebNet (ours)} & \textcolor{black}{\textbf{-0.2477} $\pm$ \textbf{0.0526}} &  \textcolor{black}{\textbf{-2.2111} $\pm$ \textbf{0.0160}} & \textcolor{black}{\textbf{-2.1043} $\pm$ \textbf{0.0766}} \\
        \bottomrule
    \end{tabular}
    \vspace{-4mm}
\end{table}

\textbf{Over-Squashing Analysis on Barbell Graphs.} To further investigate the robustness of our model to oversquashing, we use as a benchmark the barbell regression tasks introduced in \cite{bamberger2024bundle}. In this task, a model that fails to transfer any information across the single bridge edge will produce an essentially random constant and obtain a mean-squared error (MSE) close to 1; an error in the 0.4–0.6 band indicates that only a partial amount of information has overcome the bottleneck. Errors around $\approx 0.25$ and below suggest that the oversquashing has been effectively overcome. In this work, we compare Stable-ChebNet's performance on barbell graphs of varying sizes ($N=10,25,50,100$) against numerous baselines, mainly an MLP and variants of MPNNs such as GCN \cite{kipf2016semi}, GAT \cite{velivckovic2018graph}, and SAGE \cite{hamilton2017inductive}. Further description of the task is found in Appendix \ref{app:barbel_description}.

\begin{wrapfigure}{r}{0.5\linewidth}
  \centering
  \vspace{-15pt}    
  \hspace{40 pt}
  \includegraphics[width=\linewidth]{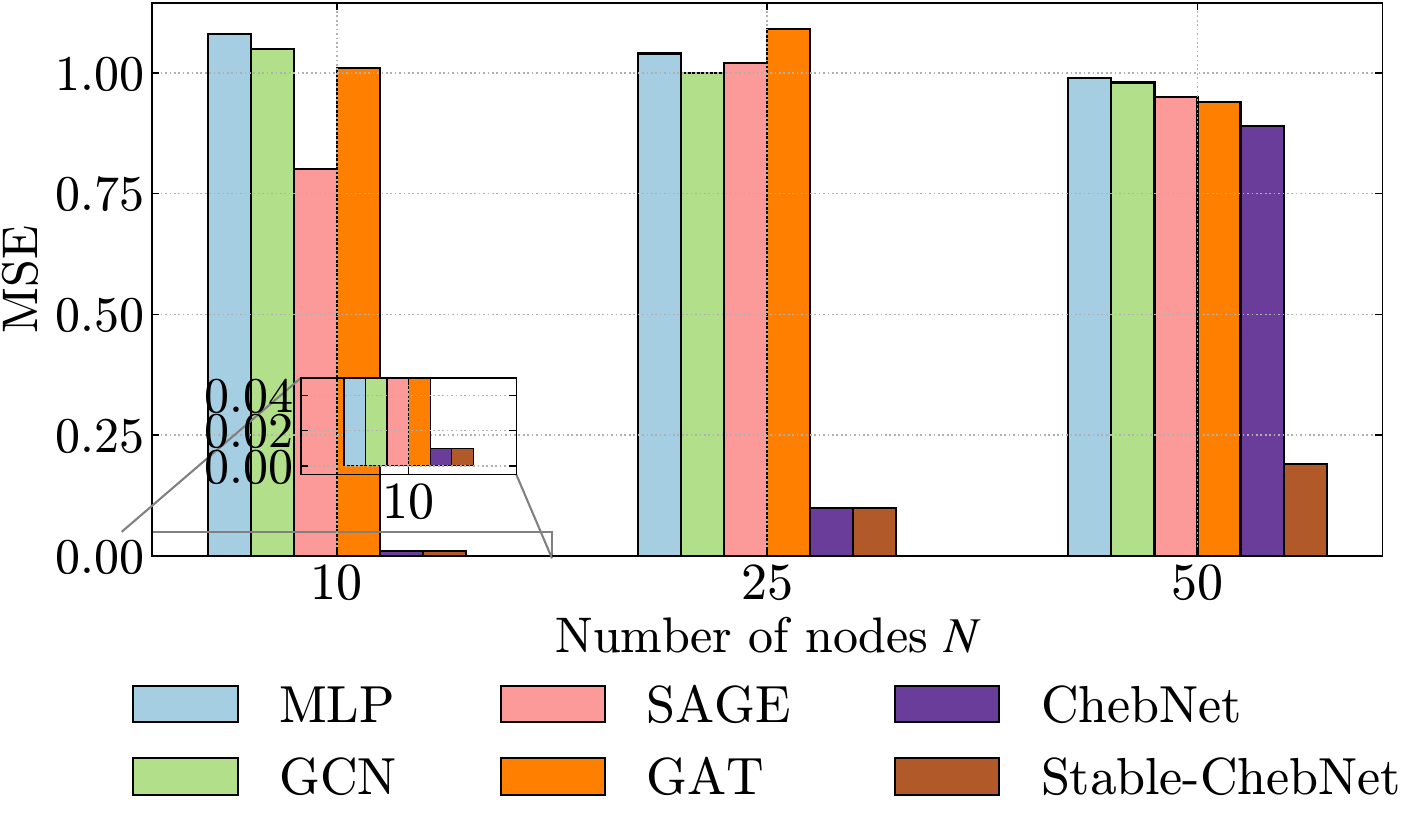}
  \caption{Mean Squared Error (MSE) comparison of various MPNN baselines at different node counts $N$ of Barbell graphs.}
  \label{fig:barbell-barplot}
  \vspace{-15pt}     
\end{wrapfigure}

\textbf{Results.} We observe in Figure \ref{fig:barbell-barplot} that both a classical ChebNet and Stable-ChebNet successfully learn the small $N = 10$ case with negligible error. However, for moderate graph sizes with $N = 50$, a classical ChebNet with fixed $K=8$ already sits in the "partial collapse" regime as its MSE increases to around 0.90 and slides towards the random-guess area as $N$ keeps growing (Table \ref{tab:combined_barbell2}). For the same range of hops K, replacing the standard update with our stable Euler-based formulation keeps the error almost two orders of magnitude smaller with an MSE below 0.20, confirming that the non-dissipative time-stepping effectively prevents the over-squashing phenomenon.

\begin{table}[h!]
\centering
\caption{Mean squared error (MSE) of ChebNet and Stable-ChebNet on the over-squashing experiment for barbell graphs. Left: sizes $N = 50, 70$ for $K = 9$ and $10$. Right: size $N = 100$ for $K = 20$.}
\label{tab:combined_barbell}

\begin{subtable}[t]{0.48\textwidth}
\centering
\begin{tabular}{llcc}
\toprule
\textbf{Method} & \textbf{K} & \textbf{50} & \textbf{70} \\
\midrule
\multirow{2}{*}{ChebNet} 
    & $K=9$ & $0.32 \pm 0.39$ & $1.08 \pm 0.05$  \\
    & $K=10$ & \textbf{0.05 $\pm$ 0.00} & $1.08 \pm 0.01$  \\
\midrule
\multirow{2}{*}{\makecell[l]{\textbf{Stable-}\\\textbf{ChebNet}}} 
    & $K=9$ & $0.17 \pm 0.11$ & $0.47 \pm 0.49$  \\
    & $K=10$ & \textbf{0.05 $\pm$ 0.00} & \textbf{0.06 $\pm$ 0.03}  \\
\bottomrule
\end{tabular}
\end{subtable}
\hfill
\begin{subtable}[t]{0.45\textwidth}
\centering
\begin{tabular}{lcc}
\toprule
\textbf{Method} & \textbf{K} & \textbf{100} \\
\midrule
ChebNet  & $K=20$ & $0.87 \pm 0.05$ \\
\multirow{2}{*}{\makecell[l]{\textbf{Stable-}\\\textbf{ChebNet (ours)}}} \\
& \textbf{$K=20$} & $\textbf{0.21} \pm \textbf{0.27}$ \\
\bottomrule
\end{tabular}
\end{subtable}
\label{tab:combined_barbell2}
\end{table}

\paragraph{Open-Graph Benchmark.}
To evaluate real-world applicability on large-scale graphs, we run experiments for node-level tasks on two large-scale graph datasets from the Open-Graph Benchmark (OGB) \cite{ogb}.
\texttt{ogbn-arxiv} is a citation network in which each node corresponds to an academic paper, and the task is node classification by predicting the subject area of unseen papers. The other dataset we use,  
\texttt{ogbn-proteins}, is a protein–protein interaction network aimed at inferring protein functions. To ensure a fair comparison and emphasize efficiency, we limit the number of parameters in our models to be within the same range as those used in existing and most recent OGB benchmarks. 

\textbf{Results.} Table \ref{table-arxiv} reports the performance on the \texttt{ogbn-arxiv} citation network, where ChebNet achieves around 73\% test accuracy, while Stable-ChebNet boosts the performance further to 75.7\%, outperforming all other methods, including a variety of MPNNs and Graph Transformers such as GraphGPS \cite{rampavsek2022recipe} and Exphormer \cite{shirzad2023exphormer}. Similarly, on the \texttt{ogbn-proteins} interaction network, ChebNet attains about 77.6\% accuracy compared to Stable-ChebNet’s 79.5\% (Table \ref{table-proteins}). Competing approaches achieve nearly 72\% for MPNN-based methods, while Transformer-based models' performances range from 77.4\% for NodeFormer \cite{wu2022nodeformer} to 79.5\% for SGFormer \cite{wu2023sgformer}. Hence, Stable-ChebNet remarkably competes and often outperforms state-of-the-art models on this benchmark, demonstrating that the Euler formulation consistently narrows the gap with and in some cases overtakes Transformer baselines such as SGFormer \cite{wu2023sgformer} and Spexphormer \cite{shirzad2024even}. Together, these results demonstrate that augmenting ChebNet with an Euler step not only addresses the classical ChebNet’s shortcomings on long-range information propagation but also performs effectively well on graphs with hundreds of thousands of nodes, in contrast to regular-sized graphs seen in previous experiments.

\begin{table}[h]
\centering
\begin{minipage}{0.48\textwidth}
\caption{Accuracy on \texttt{ogbn-arxiv}.}

\centering
\begin{tabular}{lc}
\toprule
\textbf{Model} & \textbf{ogbn-arxiv} \\
\midrule
GCN             & 71.74 $\pm$ 0.29 \\
ChebNet         & 73.27 $\pm$ 0.23 \\
ChebNetII       & 72.32 $\pm$ 0.23 \\
GraphSAGE       & 71.49 $\pm$ 0.27 \\
GAT             & 72.01 $\pm$ 0.20 \\
NodeFormer      & 59.90 $\pm$ 0.42 \\
GraphGPS        & 70.92 $\pm$ 0.04 \\
GOAT            & 72.41 $\pm$ 0.40 \\
EXPHORMER+GCN   & 72.44 $\pm$ 0.28 \\
SPEXPHORMER     & 70.82 $\pm$ 0.24 \\
\midrule
{\textbf{Stable-ChebNet (ours)}}   & \textbf{75.73 $\pm$ 0.51} \\
\bottomrule
\end{tabular}
\label{table-arxiv}
\end{minipage}
\hfill
\begin{minipage}{0.48\textwidth}
\centering
\caption{Accuracy on \texttt{ogbn-proteins}.}
\begin{tabular}{lc}
\toprule
\textbf{Model} & \textbf{ogbn-proteins} \\
\midrule
MLP            & 72.04 $\pm$ 0.48 \\
GCN            & 72.51 $\pm$ 0.35 \\
ChebNet        & 77.55 $\pm$ 0.43 \\
SGC            & 70.31 $\pm$ 0.23 \\
GCN-NSAMPLER   & 73.51 $\pm$ 1.31 \\
GAT-NSAMPLER   & 74.63 $\pm$ 1.24 \\
SIGN           & 71.24 $\pm$ 0.46 \\
NodeFormer     & 77.45 $\pm$ 1.15 \\
SGFormer       & 79.53 $\pm$ 0.38 \\
\textbf{SPEXPHORMER}    & \textbf{80.65 $\pm$ 0.07} \\
\midrule
{Stable-ChebNet (ours)}  & 79.55 $\pm$ 0.34 \\
\bottomrule
\end{tabular}
\label{table-proteins}
\end{minipage}
\end{table}

\paragraph{Long-Range Graph Benchmark (LRGB).}
LRGB \cite{dwivedi2022long} is a collection of GNN benchmarks that evaluate models on tasks involving long-range interactions. We use two of its molecular-property datasets: Peptides-func for graph classification and Peptides-struct for graph regression. 

\textbf{Results.} A detailed leaderboard is shown in Table \ref{tab:peptides}. It can be seen that Stable-ChebNet improves upon its vanilla counterpart. Together with S2GCN, it reaches an average precision (AP) above 70 on Peptide-func and a Mean Absolute Error (MAE) below 0.26 on the regression task, bearing in mind that S2GCN requires a more expensive full Laplacian eigendecomposition. Overall, our model achieves 
competitive performance on peptide structures with results competing with and often outperforming some well-known graph-based models including graph transformers such as Exphormer \cite{shirzad2023exphormer} and GraphViT \cite{he2023generalization}, state space models including Graph Mamba and GMN \cite{wang2024graph}, and rewiring methods like DRew \cite{gutteridge2023drew}. It is worth noting that the gain in AP for DRew comes at the cost of computing positional encodings (Laplacian eigenvectors) for every graph before training, while Stable-ChebNet does not use any positional encodings. 

\paragraph{Heterophilic benchmarks} We further assess Stable-ChebNet on node-classification tasks explicitly designed to stress performance under heterophily, following the standardized protocol of Platonov et al. (“Roman-empire”, “Amazon-ratings”, “Minesweeper” and “Tolokers”) \cite{platonov2023a}. We keep the exact data processing, splits, and metrics recommended therein. Concretely, we report accuracy on Roman-empire and Amazon-ratings, and ROC-AUC on Minesweeper and Tolokers averaging over four random initializations as in the protocol. 

\textbf{Link to oversmoothing, heterophily, and long-rangeness.} Our heterophilic results (Table \ref{tab:results_hetero_complete}) should not be over-interpreted as “evidence of long-range propagation” or as a direct antidote to oversmoothing. The recent position paper by Arnaiz-Rodríguez \& Errica \cite{arnaiz2025oversmoothing} argues that several widespread assumptions in the literature are often conflated: (i) that heterophily is inherently detrimental while homophily is beneficial, (ii) that long-range propagation is best evaluated on heterophilic graphs, and (iii) that performance degradation mainly arises from oversmoothing. They show that heterophily, long-range interactions, and oversmoothing are orthogonal factors: a graph may be heterophilic yet dominated by local dependencies, or homophilic yet require long-range reasoning. Hence, evaluations should focus on the nature of the learning task, not merely on global homophily ratios. In this light, our Stable-ChebNet scores on Roman-empire, Minesweeper, and Tolokers demonstrate that a stable spectral propagator has competitive performance on standardized heterophily benchmarks, but they do not by themselves certify long-rangeness. Those conclusions are better supported by our dedicated long-range tests and stability analysis. 

\begin{table*}[t]
    \centering
    \footnotesize
    \caption{Long-range benchmark results. AP is the target metric on 
    \textit{peptides-func} (higher is better), and MAE is the target metric on 
    \textit{peptides-struct} (lower is better). 
    }
    \label{tab:peptides_multirow}
    \begin{tabular}{llcc}
    \toprule
    \textbf{Model Type} & \textbf{Model} & \textbf{peptides-func (AP $\uparrow$)} & \textbf{peptides-struct (MAE $\downarrow$)} \\
    \midrule
    \multirow{7}{*}{\centering \textit{Transformer}}
     & SAN+LapPE            & 63.84 $\pm$ 1.21 & 0.2683 $\pm$ 0.0043 \\
     & TIGT            & 66.79 $\pm$ 0.74 & 0.2485 $\pm$ 0.0015 \\
     & Specformer         & 66.86 $\pm$ 0.64 & 0.2550 $\pm$ 0.0014 \\
     & Exphormer         & 65.27 $\pm$ 0.43 & 0.2481 $\pm$ 0.0007 \\
     
     & G.MLPMixer        & 69.21 $\pm$ 0.54 & 0.2475 $\pm$ 0.0015 \\
     & Graph ViT         & 69.42 $\pm$ 0.75 & 0.2449 $\pm$ 0.0016 \\
     & GRIT              & 69.88 $\pm$ 0.82 & 0.2460 $\pm$ 0.0012 \\
    \midrule
    \multirow{2}{*}{\centering \textit{Rewiring}}
     & LASER  & 64.40 $\pm$ 0.10 & 0.3043 $\pm$ 0.0019 \\   
     & DRew-GCN  & 69.96 $\pm$ 0.76 & 0.2781 $\pm$ 0.0028 \\
     & \quad +PE                & 71.50 $\pm$ 0.44 & 0.2536 $\pm$ 0.0015 \\
    \midrule
    \multirow{2}{*}{\centering \textit{State Space}}
     & Graph Mamba  & 67.39 $\pm$ 0.87 & 0.2478 $\pm$ 0.0016 \\
     & GMN  & 70.71 $\pm$ 0.83 & 0.2473 $\pm$ 0.0025 \\
     & MP-SSM & 69.93 $\pm$ 0.52 & 0.2458 $\pm$ 0.0017\\
    \midrule
    \multirow{12}{*}{\centering \textit{GNN}}
     & A-DGN      & 59.75 $\pm$ 0.44 & 0.2874 $\pm$ 0.0021 \\
     & ChebNet & 69.61 $\pm$ 0.33 & 0.2627 $\pm$ 0.0033\\             
     & ChebNetII & 68.19 $\pm$ 0.27 &  0.2618 $\pm$ 0.0058 \\             
     & GCN     & 68.60 $\pm$ 0.50 & 0.2460 $\pm$ 0.0007 \\
     & GRAMA  & 70.93 $\pm$ 0.78 & \textbf{0.2436 $\pm$ 0.0022}\\
     & GRAND           & 57.89 $\pm$ 0.62 & 0.3418 $\pm$ 0.0015 \\
     & GraphCON        & 60.22 $\pm$ 0.68 & 0.2778 $\pm$ 0.0018 \\
     & PH-DGN & 70.12 $\pm$ 0.45 & 0.2465 $\pm$ 0.0020 \\
     & SWAN & 67.51 $\pm$ 0.39 & 0.2485 $\pm$ 0.0009\\
     & PathNN        & 68.16 $\pm$ 0.26 & 0.2545 $\pm$ 0.0032 \\
     & CIN++        & 65.69 $\pm$ 1.17 & 0.2523 $\pm$ 0.0013 \\
     & \textbf{S2GCN}             & \textbf{72.75 $\pm$ 0.66} & \textbf{0.2467 $\pm$ 0.0019} \\
     & \quad +PE                 & \textbf{73.11 $\pm$ 0.66} & \textbf{0.2447 $\pm$ 0.0032} \\
     & {Stable-ChebNet (ours)}             & {70.32 $\pm$ 0.26} & {0.2542 $\pm$ 0.0030}\\
    \bottomrule
    \end{tabular}
    \label{tab:peptides}
\end{table*}

\section{Conclusion}

In this work, we have re-examined ChebNet, one of the earliest spectral GNNs, from first principles, uncovering its innate ability to capture long-range dependencies via higher-order polynomial filters, but also its susceptibility to unstable propagation dynamics as the polynomial order grows. By casting ChebNet as a continuous-time ODE and imposing antisymmetric weight constraints, we introduced Stable-ChebNet, whose forward Euler discretization yields non-dissipative, second-order–stable information flow without resorting to costly eigendecompositions, positional encodings, or graph rewiring. We provide a theoretical analysis of Stable-ChebNet showing purely imaginary Jacobian spectra and bounded layerwise sensitivity. We support the analysis with extensive experiments on a variety of synthetic and other long-range graph benchmarks. Stable-ChebNet consistently matches or outperforms state-of-the-art message-passing, rewiring, state-space, and transformer-based models , while retaining the fundamental properties of Chebyshev filters. Future work can focus on generalizing this ODE-based framework to broader spectral GNN families, whereby the stability analysis can be extended to other polynomial types and orthogonal bases.

\textbf{Impact Statement.} This work aims to advance the field of machine learning on graph-structured data which are abundant in the real world. There are many potential societal consequences of our work, none of which we feel must be specifically highlighted here.

\textbf{Code Availability} All code and datasets will be made publicly available at \href{https://github.com/ahariri13/Stable-ChebNet}{https://github.com/ahariri13/Stable-ChebNet} to support reproducibility and future work.

\begin{acksection}
AG and DB acknowledge funding from EU-EIC EMERGE (Grant No. 101070918). XD acknowledges support from the Oxford-Man Institute of Quantitative Finance and EPSRC No. EP/T023333/1.
\end{acksection}
\clearpage 

\newpage
\bibliographystyle{plain}
\bibliography{refs}

\begin{thebibliography}{10}

\bibitem{alon2021on}
Uri Alon and Eran Yahav.
\newblock On the bottleneck of graph neural networks and its practical implications.
\newblock In {\em International Conference on Learning Representations}, 2021.

\bibitem{arnaiz2025oversmoothing}
Adrian Arnaiz-Rodriguez and Federico Errica.
\newblock Oversmoothing," oversquashing", heterophily, long-range, and more: Demystifying common beliefs in graph machine learning.
\newblock {\em arXiv preprint arXiv:2505.15547}, 2025.

\bibitem{arroyo2025vanishing}
{\'A}lvaro Arroyo, Alessio Gravina, Benjamin Gutteridge, Federico Barbero, Claudio Gallicchio, Xiaowen Dong, Michael Bronstein, and Pierre Vandergheynst.
\newblock On vanishing gradients, over-smoothing, and over-squashing in gnns: Bridging recurrent and graph learning.
\newblock {\em arXiv preprint arXiv:2502.10818}, 2025.

\bibitem{attali2024rewiringtechniquesmitigateoversquashing}
Hugo Attali, Davide Buscaldi, and Nathalie Pernelle.
\newblock Rewiring techniques to mitigate oversquashing and oversmoothing in gnns: A survey.
\newblock {\em arXiv preprint arXiv:2411.17429}, 2024.

\bibitem{bamberger2024bundle}
Jacob Bamberger, Federico Barbero, Xiaowen Dong, and Michael Bronstein.
\newblock Bundle neural networks for message diffusion on graphs.
\newblock {\em arXiv preprint arXiv:2405.15540}, 2024.

\bibitem{bamberger2025measuring}
Jacob Bamberger, Benjamin Gutteridge, Scott~le Roux, Michael~M Bronstein, and Xiaowen Dong.
\newblock On measuring long-range interactions in graph neural networks.
\newblock {\em arXiv preprint arXiv:2506.05971}, 2025.

\bibitem{barbero2023locality}
Federico Barbero, Ameya Velingker, Amin Saberi, Michael~M. Bronstein, and Francesco~Di Giovanni.
\newblock Locality-aware graph rewiring in {GNN}s.
\newblock In {\em The Twelfth International Conference on Learning Representations}, 2024.

\bibitem{behrouz2024graph}
Ali Behrouz and Farnoosh Hashemi.
\newblock Graph mamba: Towards learning on graphs with state space models.
\newblock In {\em Proceedings of the 30th ACM SIGKDD Conference on Knowledge Discovery and Data Mining}, pages 119--130, 2024.

\bibitem{black2023understanding}
Mitchell Black, Zhengchao Wan, Amir Nayyeri, and Yusu Wang.
\newblock Understanding oversquashing in gnns through the lens of effective resistance.
\newblock In {\em International Conference on Machine Learning}, pages 2528--2547. PMLR, 2023.

\bibitem{bo2023specformer}
Deyu Bo, Chuan Shi, Lele Wang, and Renjie Liao.
\newblock Specformer: Spectral graph neural networks meet transformers.
\newblock {\em arXiv preprint arXiv:2303.01028}, 2023.

\bibitem{FAGCN}
Deyu Bo, Xiao Wang, Chuan Shi, and Huawei Shen.
\newblock Beyond low-frequency information in graph convolutional networks.
\newblock {\em Proceedings of the AAAI Conference on Artificial Intelligence}, 35(5):3950--3957, May 2021.

\bibitem{bodnar2022neural}
Cristian Bodnar, Francesco~Di Giovanni, Benjamin~P. Chamberlain, Pietro Liò, and Michael~M. Bronstein.
\newblock Neural sheaf diffusion: A topological perspective on heterophily and oversmoothing in {GNN}s, 2022.

\bibitem{gatedgcn}
Xavier Bresson and Thomas Laurent.
\newblock {Residual Gated Graph ConvNets}.
\newblock {\em arXiv preprint arXiv:1711.07553}, 2018.

\bibitem{bruna2013spectral}
Joan Bruna, Wojciech Zaremba, Arthur Szlam, and Yann LeCun.
\newblock Spectral networks and locally connected networks on graphs.
\newblock {\em arXiv preprint arXiv:1312.6203}, 2013.

\bibitem{cai2020note}
Chen Cai and Yusu Wang.
\newblock A note on over-smoothing for graph neural networks.
\newblock {\em arXiv preprint arXiv:2006.13318}, 2020.

\bibitem{mpssm}
Andrea Ceni, Alessio Gravina, Claudio Gallicchio, Davide Bacciu, Carola-Bibiane Schonlieb, and Moshe Eliasof.
\newblock {Message-Passing State-Space Models: Improving Graph Learning with Modern Sequence Modeling}.
\newblock {\em arXiv preprint arXiv:2505.18728}, 2025.

\bibitem{chamberlain2021grand}
Benjamin~Paul Chamberlain, James Rowbottom, Maria Gorinova, Stefan Webb, Emanuele Rossi, and Michael~M Bronstein.
\newblock {GRAND}: {Graph} neural diffusion.
\newblock In {\em International Conference on Machine Learning (ICML)}, pages 1407--1418. PMLR, 2021.

\bibitem{chang2018antisymmetricrnn}
Bo~Chang, Minmin Chen, Eldad Haber, and Ed~H. Chi.
\newblock Antisymmetric{RNN}: A dynamical system view on recurrent neural networks.
\newblock In {\em International Conference on Learning Representations}, 2019.

\bibitem{nagphormer}
Jinsong Chen, Kaiyuan Gao, Gaichao Li, and Kun He.
\newblock Nagphormer: A tokenized graph transformer for node classification in large graphs.
\newblock In {\em Proceedings of the International Conference on Learning Representations}, 2023.

\bibitem{gcnii}
Ming Chen, Zhewei Wei, Zengfeng Huang, Bolin Ding, and Yaliang Li.
\newblock {Simple and Deep Graph Convolutional Networks}.
\newblock In Hal~Daumé III and Aarti Singh, editors, {\em Proceedings of the 37th International Conference on Machine Learning}, volume 119 of {\em Proceedings of Machine Learning Research}, pages 1725--1735. PMLR, 13--18 Jul 2020.

\bibitem{GPR_GNN}
Eli Chien, Jianhao Peng, Pan Li, and Olgica Milenkovic.
\newblock Adaptive universal generalized pagerank graph neural network.
\newblock In {\em International Conference on Learning Representations}, 2021.

\bibitem{choi2025topologyinformedgraphtransformer}
Yun~Young Choi, Sun~Woo Park, Minho Lee, and Youngho Woo.
\newblock Topology-informed graph transformer.
\newblock {\em arXiv preprint arXiv:2402.02005}, 2025.

\bibitem{Choromanski2020}
Krzysztof Choromanski, Marcin Kuczynski, Jacek Cieszkowski, Paul~L. Beletsky, Konrad~M. Smith, Wojciech Gajewski, Gabriel~De Masson, Tomasz~Z. Broniatowski, Antonina~B. Gorny, Leszek~M. Kaczmarek, and Stanislaw~K. Andrzejewski.
\newblock Performers: A new approach to scaling transformers.
\newblock {\em Proceedings of the 37th International Conference on Machine Learning (ICML)}, pages 2020--2031, 2020.

\bibitem{corso2020principal}
Gabriele Corso, Luca Cavalleri, Dominique Beaini, Pietro Li{\`o}, and Petar Veli{\v{c}}kovi{\'c}.
\newblock Principal neighbourhood aggregation for graph nets.
\newblock In {\em Advances in Neural Information Processing Systems (NeurIPS)}, volume~33, pages 13260--13271, 2020.

\bibitem{defferrard2016convolutional}
Micha{\"e}l Defferrard, Xavier Bresson, and Pierre Vandergheynst.
\newblock Convolutional neural networks on graphs with fast localized spectral filtering.
\newblock {\em Advances in neural information processing systems}, 29, 2016.

\bibitem{deng2024polynormer}
Chenhui Deng, Zichao Yue, and Zhiru Zhang.
\newblock Polynormer: Polynomial-expressive graph transformer in linear time.
\newblock In {\em The Twelfth International Conference on Learning Representations}, 2024.

\bibitem{di2023over}
Francesco Di~Giovanni, Lorenzo Giusti, Federico Barbero, Giulia Luise, Pietro Lio, and Michael~M Bronstein.
\newblock On over-squashing in message passing neural networks: The impact of width, depth, and topology.
\newblock In {\em International conference on machine learning}, pages 7865--7885. PMLR, 2023.

\bibitem{di2022graph}
Francesco Di~Giovanni, James Rowbottom, Benjamin~P Chamberlain, Thomas Markovich, and Michael~M Bronstein.
\newblock Graph neural networks as gradient flows: understanding graph convolutions via energy.
\newblock {\em arXiv preprint arXiv:2206.10991}, 2022.

\bibitem{GBK_GNN}
Lun Du, Xiaozhou Shi, Qiang Fu, Xiaojun Ma, Hengyu Liu, Shi Han, and Dongmei Zhang.
\newblock Gbk-gnn: Gated bi-kernel graph neural networks for modeling both homophily and heterophily.
\newblock In {\em Proceedings of the ACM Web Conference 2022}, WWW '22, page 1550–1558, New York, NY, USA, 2022. Association for Computing Machinery.

\bibitem{dwivedi2020generalization}
Vijay~Prakash Dwivedi and Xavier Bresson.
\newblock A generalization of transformer networks to graphs.
\newblock {\em arXiv preprint arXiv:2012.09699}, 2020.

\bibitem{gnnbenchmark}
Vijay~Prakash Dwivedi, Chaitanya~K. Joshi, Anh~Tuan Luu, Thomas Laurent, Yoshua Bengio, and Xavier Bresson.
\newblock Benchmarking graph neural networks.
\newblock {\em J. Mach. Learn. Res.}, 24(1), January 2023.

\bibitem{dwivedi2022graph}
Vijay~Prakash Dwivedi, Anh~Tuan Luu, Thomas Laurent, Yoshua Bengio, and Xavier Bresson.
\newblock Graph neural networks with learnable structural and positional representations.
\newblock In {\em International Conference on Learning Representations}, 2022.

\bibitem{dwivedi2022long}
Vijay~Prakash Dwivedi, Ladislav Ramp{\'a}{\v{s}}ek, Michael Galkin, Ali Parviz, Guy Wolf, Anh~Tuan Luu, and Dominique Beaini.
\newblock Long range graph benchmark.
\newblock {\em Advances in Neural Information Processing Systems}, 35:22326--22340, 2022.

\bibitem{grama}
Moshe Eliasof, Alessio Gravina, Andrea Ceni, Claudio Gallicchio, Davide Bacciu, and Carola-Bibiane Schönlieb.
\newblock {Graph Adaptive Autoregressive Moving Average Models}.
\newblock In {\em Forty-second International Conference on Machine Learning}, 2025.

\bibitem{amp}
Federico Errica, Henrik Christiansen, Viktor Zaverkin, Takashi Maruyama, Mathias Niepert, and Francesco Alesiani.
\newblock Adaptive message passing: A general framework to mitigate oversmoothing, oversquashing, and underreaching.
\newblock In {\em Proceedings of the 42nd International Conference on Machine Learning}, volume 267 of {\em Proceedings of Machine Learning Research}, pages 15490--15515. PMLR, 13--19 Jul 2025.

\bibitem{pmlr-v231-fesser24a}
Lukas Fesser and Melanie Weber.
\newblock Mitigating over-smoothing and over-squashing using augmentations of forman-ricci curvature.
\newblock In Soledad Villar and Benjamin Chamberlain, editors, {\em Proceedings of the Second Learning on Graphs Conference}, volume 231 of {\em Proceedings of Machine Learning Research}, pages 19:1--19:28. PMLR, 27--30 Nov 2024.

\bibitem{finkelshtein2024cooperative}
Ben Finkelshtein, Xingyue Huang, Michael~M. Bronstein, and Ismail~Ilkan Ceylan.
\newblock {Cooperative Graph Neural Networks}.
\newblock In {\em Forty-first International Conference on Machine Learning}, 2024.

\bibitem{frasca2020signscalableinceptiongraph}
Fabrizio Frasca, Emanuele Rossi, Davide Eynard, Ben Chamberlain, Michael Bronstein, and Federico Monti.
\newblock Sign: Scalable inception graph neural networks.
\newblock {\em arXiv preprint arXiv:2004.11198}, 2020.

\bibitem{DIGL}
Johannes Gasteiger, Stefan Wei\ss~enberger, and Stephan G\"{u}nnemann.
\newblock {Diffusion Improves Graph Learning}.
\newblock In {\em Advances in Neural Information Processing Systems}, volume~32. Curran Associates, Inc., 2019.

\bibitem{geisler2024spatio}
Simon~Markus Geisler, Arthur Kosmala, Daniel Herbst, and Stephan G{\"u}nnemann.
\newblock Spatio-spectral graph neural networks.
\newblock {\em Advances in Neural Information Processing Systems}, 37:49022--49080, 2024.

\bibitem{gilmer2017neural}
Justin Gilmer, Samuel~S Schoenholz, Patrick~F Riley, Oriol Vinyals, and George~E Dahl.
\newblock Neural message passing for quantum chemistry.
\newblock In {\em International conference on machine learning}, pages 1263--1272. PMLR, 2017.

\bibitem{giusti2023cinenhancingtopologicalmessage}
Lorenzo Giusti, Teodora Reu, Francesco Ceccarelli, Cristian Bodnar, and Pietro Liò.
\newblock Cin++: Enhancing topological message passing.
\newblock {\em arXiv preprint arXiv:2306.03561}, 2023.

\bibitem{gori2005new}
M~Gori, G~Monfardini, and F~Scarselli.
\newblock A new model for learning in graph domains.
\newblock In {\em Proceedings. 2005 IEEE International Joint Conference on Neural Networks, 2005.}, volume~2, pages 729--734. IEEE, 2005.

\bibitem{gravina_dynamic_survey}
Alessio Gravina and Davide Bacciu.
\newblock {Deep Learning for Dynamic Graphs: Models and Benchmarks}.
\newblock {\em IEEE Transactions on Neural Networks and Learning Systems}, 35(9):11788--11801, 2024.

\bibitem{gravina_adgn}
Alessio Gravina, Davide Bacciu, and Claudio Gallicchio.
\newblock {Anti-Symmetric DGN: a stable architecture for Deep Graph Networks}.
\newblock In {\em The Eleventh International Conference on Learning Representations}, 2023.

\bibitem{gravina_swan}
Alessio Gravina, Moshe Eliasof, Claudio Gallicchio, Davide Bacciu, and Carola-Bibiane Sch\"onlieb.
\newblock On oversquashing in graph neural networks through the lens of dynamical systems.
\newblock In {\em The 39th Annual AAAI Conference on Artificial Intelligence}, 2025.

\bibitem{gravina_randomized_adgn}
Alessio Gravina, Claudio Gallicchio, and Davide Bacciu.
\newblock {Non-dissipative Propagation by Randomized Anti-symmetric Deep Graph Networks}.
\newblock In {\em Machine Learning and Principles and Practice of Knowledge Discovery in Databases}, pages 25--36, Cham, 2025. Springer Nature Switzerland.

\bibitem{gravina_ctan}
Alessio Gravina, Giulio Lovisotto, Claudio Gallicchio, Davide Bacciu, and Claas Grohnfeldt.
\newblock {Long Range Propagation on Continuous-Time Dynamic Graphs}.
\newblock In {\em Proceedings of the 41st International Conference on Machine Learning}, volume 235 of {\em Proceedings of Machine Learning Research}, pages 16206--16225. PMLR, 21--27 Jul 2024.

\bibitem{gutteridge2023drew}
Benjamin Gutteridge, Xiaowen Dong, Michael~M Bronstein, and Francesco Di~Giovanni.
\newblock Drew: Dynamically rewired message passing with delay.
\newblock In {\em International Conference on Machine Learning}, pages 12252--12267. PMLR, 2023.

\bibitem{HaberRuthotto2017}
E.~Haber and L.~Ruthotto.
\newblock Stable architectures for deep neural networks.
\newblock {\em Inverse Problems}, 34(1), 2017.

\bibitem{hamilton2017inductive}
William~L Hamilton, Rex Ying, and Jure Leskovec.
\newblock Inductive representation learning on large graphs.
\newblock In {\em Advances in Neural Information Processing Systems (NeurIPS)}, volume~30, 2017.

\bibitem{he2022convolutional}
Mingguo He, Zhewei Wei, and Ji-Rong Wen.
\newblock Convolutional neural networks on graphs with chebyshev approximation, revisited.
\newblock {\em Advances in Neural Information Processing Systems}, 35:7264--7276, 2022.

\bibitem{he2023generalization}
Xiaoxin He, Bryan Hooi, Thomas Laurent, Adam Perold, Yann LeCun, and Xavier Bresson.
\newblock A generalization of vit/mlp-mixer to graphs.
\newblock In {\em International conference on machine learning}, pages 12724--12745. PMLR, 2023.

\bibitem{gravina_phdgn}
Simon Heilig, Alessio Gravina, Alessandro Trenta, Claudio Gallicchio, and Davide Bacciu.
\newblock {Port-Hamiltonian Architectural Bias for Long-Range Propagation in Deep Graph Networks}.
\newblock In {\em The Thirteenth International Conference on Learning Representations}, 2025.

\bibitem{ogb}
Weihua Hu, Matthias Fey, Marinka Zitnik, Yuxiao Dong, Hongyu Ren, Bowen Liu, Michele Catasta, and Jure Leskovec.
\newblock Open graph benchmark: datasets for machine learning on graphs.
\newblock In {\em Proceedings of the 34th International Conference on Neural Information Processing Systems}, NIPS '20. Curran Associates Inc., 2020.

\bibitem{karhadkar2022fosr}
Kedar Karhadkar, Pradeep~Kr. Banerjee, and Guido Mont{\'{u}}far.
\newblock Fosr: First-order spectral rewiring for addressing oversquashing in gnns.
\newblock In {\em The Eleventh International Conference on Learning Representations, {ICLR}}, 2023.

\bibitem{kipf2016semi}
Thomas~N Kipf and Max Welling.
\newblock Semi-supervised classification with graph convolutional networks.
\newblock {\em International Conference on Learning Representations (ICLR)}, 2017.

\bibitem{goat}
Kezhi Kong, Jiuhai Chen, John Kirchenbauer, Renkun Ni, C.~Bayan Bruss, and Tom Goldstein.
\newblock {GOAT}: A global transformer on large-scale graphs.
\newblock In Andreas Krause, Emma Brunskill, Kyunghyun Cho, Barbara Engelhardt, Sivan Sabato, and Jonathan Scarlett, editors, {\em Proceedings of the 40th International Conference on Machine Learning}, volume 202 of {\em Proceedings of Machine Learning Research}, pages 17375--17390. PMLR, 23--29 Jul 2023.

\bibitem{san}
Devin Kreuzer, Dominique Beaini, Will Hamilton, Vincent L{\'e}tourneau, and Prudencio Tossou.
\newblock Rethinking graph transformers with spectral attention.
\newblock {\em Advances in Neural Information Processing Systems}, 34:21618--21629, 2021.

\bibitem{GloGNN}
Xiang Li, Renyu Zhu, Yao Cheng, Caihua Shan, Siqiang Luo, Dongsheng Li, and Weining Qian.
\newblock Finding global homophily in graph neural networks when meeting heterophily.
\newblock In Kamalika Chaudhuri, Stefanie Jegelka, Le~Song, Csaba Szepesvari, Gang Niu, and Sivan Sabato, editors, {\em Proceedings of the 39th International Conference on Machine Learning}, volume 162 of {\em Proceedings of Machine Learning Research}, pages 13242--13256. PMLR, 17--23 Jul 2022.

\bibitem{liang2025towards}
Huidong Liang, Haitz S{\'a}ez de~Oc{\'a}riz Borde, Baskaran Sripathmanathan, Michael Bronstein, and Xiaowen Dong.
\newblock Towards quantifying long-range interactions in graph machine learning: a large graph dataset and a measurement.
\newblock {\em arXiv preprint arXiv:2503.09008}, 2025.

\bibitem{luan2024heterophilicgraphlearninghandbook}
Sitao Luan, Chenqing Hua, Qincheng Lu, Liheng Ma, Lirong Wu, Xinyu Wang, Minkai Xu, Xiao-Wen Chang, Doina Precup, Rex Ying, Stan~Z. Li, Jian Tang, Guy Wolf, and Stefanie Jegelka.
\newblock The heterophilic graph learning handbook: Benchmarks, models, theoretical analysis, applications and challenges, 2024.

\bibitem{grit}
Liheng Ma, Chen Lin, Derek Lim, Adriana Romero-Soriano, Puneet~K. Dokania, Mark Coates, Philip Torr, and Ser-Nam Lim.
\newblock Graph inductive biases in transformers without message passing.
\newblock In {\em Proceedings of the 40th International Conference on Machine Learning}, volume 202 of {\em Proceedings of Machine Learning Research}, pages 23321--23337. PMLR, 23--29 Jul 2023.

\bibitem{maskey2024fractional}
Sohir Maskey, Raffaele Paolino, Aras Bacho, and Gitta Kutyniok.
\newblock A fractional graph laplacian approach to oversmoothing.
\newblock {\em Advances in Neural Information Processing Systems}, 36, 2024.

\bibitem{FSGNN}
Sunil~Kumar Maurya, Xin Liu, and Tsuyoshi Murata.
\newblock Simplifying approach to node classification in graph neural networks.
\newblock {\em Journal of Computational Science}, 62:101695, 2022.

\bibitem{pathnn}
Gaspard Michel, Giannis Nikolentzos, Johannes Lutzeyer, and Michalis Vazirgiannis.
\newblock Path neural networks: expressive and accurate graph neural networks.
\newblock In {\em Proceedings of the 40th International Conference on Machine Learning}, ICML'23. JMLR.org, 2023.

\bibitem{NN4G}
Alessio Micheli.
\newblock {Neural Network for Graphs: A Contextual Constructive Approach}.
\newblock {\em IEEE Transactions on Neural Networks}, 20(3):498--511, 2009.

\bibitem{borf}
Khang Nguyen, Hieu Nong, Vinh Nguyen, Nhat Ho, Stanley Osher, and Tan Nguyen.
\newblock Revisiting over-smoothing and over-squashing using ollivier-ricci curvature.
\newblock In {\em Proceedings of the 40th International Conference on Machine Learning}, ICML'23. JMLR.org, 2023.

\bibitem{oono2020graph}
Kenta Oono and Taiji Suzuki.
\newblock Graph neural networks exponentially lose expressive power for node classification.
\newblock In {\em International Conference on Learning Representations}, 2020.

\bibitem{platonov2023a}
Oleg Platonov, Denis Kuznedelev, Michael Diskin, Artem Babenko, and Liudmila Prokhorenkova.
\newblock A critical look at the evaluation of {GNN}s under heterophily: Are we really making progress?
\newblock In {\em The Eleventh International Conference on Learning Representations}, 2023.

\bibitem{rampavsek2022recipe}
Ladislav Ramp{\'a}{\v{s}}ek, Michael Galkin, Vijay~Prakash Dwivedi, Anh~Tuan Luu, Guy Wolf, and Dominique Beaini.
\newblock Recipe for a general, powerful, scalable graph transformer.
\newblock {\em Advances in Neural Information Processing Systems}, 35:14501--14515, 2022.

\bibitem{rusch2023survey}
T.~Konstantin Rusch, Michael~M. Bronstein, and Siddhartha Mishra.
\newblock {A Survey on Oversmoothing in Graph Neural Networks}.
\newblock {\em arXiv preprint arXiv:2303.10993}, 2023.

\bibitem{rusch2022graph}
T~Konstantin Rusch, Ben Chamberlain, James Rowbottom, Siddhartha Mishra, and Michael Bronstein.
\newblock Graph-coupled oscillator networks.
\newblock In {\em International Conference on Machine Learning}, pages 18888--18909. PMLR, 2022.

\bibitem{scarselli2008graph}
Franco Scarselli, Marco Gori, Ah~Chung Tsoi, Markus Hagenbuchner, and Gabriele Monfardini.
\newblock The graph neural network model.
\newblock {\em IEEE transactions on neural networks}, 20(1):61--80, 2008.

\bibitem{sestak2024vn}
Florian Sestak, Lisa Schneckenreiter, Johannes Brandstetter, Sepp Hochreiter, Andreas Mayr, and G{\"u}nter Klambauer.
\newblock Vn-egnn: E (3)-equivariant graph neural networks with virtual nodes enhance protein binding site identification.
\newblock {\em arXiv preprint arXiv:2404.07194}, 2024.

\bibitem{shehzad2024graphtransformerssurvey}
Ahsan Shehzad, Feng Xia, Shagufta Abid, Ciyuan Peng, Shuo Yu, Dongyu Zhang, and Karin Verspoor.
\newblock Graph transformers: A survey, 2024.

\bibitem{shi2023exposition}
Dai Shi, Andi Han, Lequan Lin, Yi~Guo, and Junbin Gao.
\newblock {Exposition on over-squashing problem on GNNs: Current Methods, Benchmarks and Challenges}, 2023.

\bibitem{ijcai2021p0214}
Yunsheng Shi, Zhengjie Huang, Shikun Feng, Hui Zhong, Wenjing Wang, and Yu~Sun.
\newblock Masked label prediction: Unified message passing model for semi-supervised classification.
\newblock In Zhi-Hua Zhou, editor, {\em Proceedings of the Thirtieth International Joint Conference on Artificial Intelligence, {IJCAI-21}}, pages 1548--1554. International Joint Conferences on Artificial Intelligence Organization, 8 2021.
\newblock Main Track.

\bibitem{shirzad2024even}
Hamed Shirzad, Honghao Lin, Balaji Venkatachalam, Ameya Velingker, David~P Woodruff, and Danica~J Sutherland.
\newblock Even sparser graph transformers.
\newblock {\em Advances in Neural Information Processing Systems}, 37:71277--71305, 2024.

\bibitem{shirzad2023exphormer}
Hamed Shirzad, Ameya Velingker, Balaji Venkatachalam, Danica~J Sutherland, and Ali~Kemal Sinop.
\newblock Exphormer: Sparse transformers for graphs.
\newblock In {\em International Conference on Machine Learning}, pages 31613--31632. PMLR, 2023.

\bibitem{sperduti1993encoding}
Alessandro Sperduti.
\newblock Encoding labeled graphs by labeling raam.
\newblock {\em Advances in Neural Information Processing Systems}, 6, 1993.

\bibitem{topping2022understanding}
Jake Topping, Francesco~Di Giovanni, Benjamin~Paul Chamberlain, Xiaowen Dong, and Michael~M. Bronstein.
\newblock Understanding over-squashing and bottlenecks on graphs via curvature.
\newblock In {\em International Conference on Learning Representations}, 2022.

\bibitem{velivckovic2018graph}
Petar Veli{\v{c}}kovi{\'c}, Guillem Cucurull, Arantxa Casanova, Adriana Romero, Pietro Lio, and Yoshua Bengio.
\newblock Graph attention networks.
\newblock In {\em Proceedings of the International Conference on Learning Representations (ICLR)}, 2018.

\bibitem{wang2024graph}
Chloe Wang, Oleksii Tsepa, Jun Ma, and Bo~Wang.
\newblock Graph-mamba: Towards long-range graph sequence modeling with selective state spaces.
\newblock {\em arXiv preprint arXiv:2402.00789}, 2024.

\bibitem{jacobiconv}
Xiyuan Wang and Muhan Zhang.
\newblock How powerful are spectral graph neural networks.
\newblock In Kamalika Chaudhuri, Stefanie Jegelka, Le~Song, Csaba Szepesvari, Gang Niu, and Sivan Sabato, editors, {\em Proceedings of the 39th International Conference on Machine Learning}, volume 162 of {\em Proceedings of Machine Learning Research}, pages 23341--23362. PMLR, 17--23 Jul 2022.

\bibitem{DGC}
Yifei Wang, Yisen Wang, Jiansheng Yang, and Zhouchen Lin.
\newblock {Dissecting the Diffusion Process in Linear Graph Convolutional Networks}.
\newblock In {\em Advances in Neural Information Processing Systems}, volume~34, pages 5758--5769. Curran Associates, Inc., 2021.

\bibitem{SGC}
Felix Wu, Amauri Souza, Tianyi Zhang, Christopher Fifty, Tao Yu, and Kilian Weinberger.
\newblock {Simplifying Graph Convolutional Networks}.
\newblock In Kamalika Chaudhuri and Ruslan Salakhutdinov, editors, {\em Proceedings of the 36th International Conference on Machine Learning}, volume~97 of {\em Proceedings of Machine Learning Research}, pages 6861--6871. PMLR, 09--15 Jun 2019.

\bibitem{wu2022nodeformer}
Qitian Wu, Wentao Zhao, Zenan Li, David~P Wipf, and Junchi Yan.
\newblock Nodeformer: A scalable graph structure learning transformer for node classification.
\newblock {\em Advances in Neural Information Processing Systems}, 35:27387--27401, 2022.

\bibitem{wu2023sgformer}
Qitian Wu, Wentao Zhao, Chenxiao Yang, Hengrui Zhang, Fan Nie, Haitian Jiang, Yatao Bian, and Junchi Yan.
\newblock Sgformer: Simplifying and empowering transformers for large-graph representations.
\newblock {\em Advances in Neural Information Processing Systems}, 36:64753--64773, 2023.

\bibitem{xu2019powerful}
Keyulu Xu, Weihua Hu, Jure Leskovec, and Stefanie Jegelka.
\newblock How powerful are graph neural networks?
\newblock In {\em International Conference on Learning Representations (ICLR)}, 2019.

\bibitem{graphormer}
Chengxuan Ying, Tianle Cai, Shengjie Luo, Shuxin Zheng, Guolin Ke, Di~He, Yanming Shen, and Tie-Yan Liu.
\newblock Do transformers really perform bad for graph representation?
\newblock In {\em Proceedings of the 35th International Conference on Neural Information Processing Systems}, NIPS '21. Curran Associates Inc., 2021.

\bibitem{Zaheer2020}
Manzil Zaheer, Guru prasad G.~H., Lihong Wang, S.~V. K. N.~L. Wang, Yujia Li, Jakub Konečný, Shalmali Joshi, Danqi Chen, Jennifer~R. R., Zhenyu Zhang, Shalini Devaraj, and Srinivas Narayanan.
\newblock Bigbird: Transformers for longer sequences.
\newblock {\em Proceedings of the 37th International Conference on Machine Learning (ICML)}, pages 12168--12178, 2020.

\bibitem{zeng2019graphsaint}
Hanqing Zeng, Hongkuan Zhou, Ajitesh Srivastava, Rajgopal Kannan, and Viktor Prasanna.
\newblock Graphsaint: Graph sampling based inductive learning method.
\newblock {\em arXiv preprint arXiv:1907.04931}, 2019.

\bibitem{CPGNN}
Jiong Zhu, Ryan~A. Rossi, Anup Rao, Tung Mai, Nedim Lipka, Nesreen~K. Ahmed, and Danai Koutra.
\newblock Graph neural networks with heterophily.
\newblock {\em Proceedings of the AAAI Conference on Artificial Intelligence}, 35(12):11168--11176, May 2021.

\bibitem{h2gcn}
Jiong Zhu, Yujun Yan, Lingxiao Zhao, Mark Heimann, Leman Akoglu, and Danai Koutra.
\newblock Beyond homophily in graph neural networks: Current limitations and effective designs.
\newblock In H.~Larochelle, M.~Ranzato, R.~Hadsell, M.F. Balcan, and H.~Lin, editors, {\em Advances in Neural Information Processing Systems}, volume~33, pages 7793--7804. Curran Associates, Inc., 2020.

\end{thebibliography}

\newpage
\section*{NeurIPS Paper Checklist}

\begin{enumerate}

\item {\bf Claims}
    \item[] Question: Do the main claims made in the abstract and introduction accurately reflect the paper's contributions and scope?
    \item[] Answer: \answerYes{} 
    \item[] Justification: All the claims found in the abstract and introduction are supported by the results and theorems discussed in the paper.

\item {\bf Limitations}
    \item[] Question: Does the paper discuss the limitations of the work performed by the authors?
    \item[] Answer: \answerYes{} 
    \item[] Justification: We mention the scope of this work that is limited to ChebNet and leave the expansion to other Spectral GNN methods for future work.

\item {\bf Theory Assumptions and Proofs}
    \item[] Question: For each theoretical result, does the paper provide the full set of assumptions and a complete (and correct) proof?
    \item[] Answer: \answerYes{} 
    \item[] Justification: The paper provides a set of theorems that justify and motivate the choice of the developed model. The corresponding proofs are also presented and can be found in the appendices.

    \item {\bf Experimental Result Reproducibility}
    \item[] Question: Does the paper fully disclose all the information needed to reproduce the main experimental results of the paper to the extent that it affects the main claims and/or conclusions of the paper (regardless of whether the code and data are provided or not)?
    \item[] Answer: \answerYes{} 
    \item[] Justification: The datasets we use are publicly available and we share the code in the supplementary materials folder. The set of sweeped hyperparameters to reproduce the results is also shared. 

\item {\bf Open access to data and code}
    \item[] Question: Does the paper provide open access to the data and code, with sufficient instructions to faithfully reproduce the main experimental results, as described in supplemental material?
    \item[] Answer: \answerYes{}
    \item[] Justification:  The provided repository includes all the datasets used as well as the necessary code to reproduce the experimental results.

\item {\bf Experimental Setting/Details}
    \item[] Question: Does the paper specify all the training and test details (e.g., data splits, hyperparameters, how they were chosen, type of optimizer, etc.) necessary to understand the results?
    \item[] Answer: \answerYes{} 
    \item[] Justification: We specify that we follow the same training procedure for the datasets as their papers of reference. We also show our hyperparameter sweep in the Appendix.

\item {\bf Experiment Statistical Significance}
    \item[] Question: Does the paper report error bars suitably and correctly defined or other appropriate information about the statistical significance of the experiments?
    \item[] Answer: \answerYes{} 
    \item[] Justification: We provide all our results in terms of graph performance in the format ``mean $\pm$ standard error of the mean".

\item {\bf Experiments Compute Resources}
    \item[] Question: For each experiment, does the paper provide sufficient information on the computer resources (type of compute workers, memory, time of execution) needed to reproduce the experiments?
    \item[] Answer: \answerYes{} 
    \item[] Justification: We provide detailed information about the used resources in the Appendix.
    
\item {\bf Code Of Ethics}
    \item[] Question: Does the research conducted in the paper conform, in every respect, with the NeurIPS Code of Ethics \url{https://neurips.cc/public/EthicsGuidelines}?
    \item[] Answer: \answerYes{} 
    \item[] Justification: We preserve anonymity in our submission and our submission abides by the NeurIPS Code of Ethics.

\item {\bf Broader Impacts}
    \item[] Question: Does the paper discuss both potential positive societal impacts and negative societal impacts of the work performed?
    \item[] Answer: \answerYes{} 
    \item[] Justification: We discuss potential positive and negative societal impacts after the conclusion section.
    
\item {\bf Safeguards}
    \item[] Question: Does the paper describe safeguards that have been put in place for responsible release of data or models that have a high risk for misuse (e.g., pretrained language models, image generators, or scraped datasets)?
    \item[] Answer: \answerNA{} 
    \item[] Justification: The release of our data and models does not pose a direct risk of misuse.

\item {\bf Licenses for existing assets}
    \item[] Question: Are the creators or original owners of assets (e.g., code, data, models), used in the paper, properly credited and are the license and terms of use explicitly mentioned and properly respected?
    \item[] Answer: \answerYes{} 
    \item[] Justification: We provide citations for all sources of code and/or data used, both in the paper and in the accompanying repository. For the datasets, we reference the original open-source work they are based on, in full accordance with their licensing terms.

\item {\bf New Assets}
    \item[] Question: Are new assets introduced in the paper well documented and is the documentation provided alongside the assets?
    \item[] Answer: \answerNA{} 
    \item[] Justification: We do not release new datasets.

\item {\bf Crowdsourcing and Research with Human Subjects}
    \item[] Question: For crowdsourcing experiments and research with human subjects, does the paper include the full text of instructions given to participants and screenshots, if applicable, as well as details about compensation (if any)? 
    \item[] Answer: \answerNA{} 
    \item[] Justification: No crowdsourcing or human subjects were involved in the experiments conducted for this paper.

\item {\bf Institutional Review Board (IRB) Approvals or Equivalent for Research with Human Subjects}
    \item[] Question: Does the paper describe potential risks incurred by study participants, whether such risks were disclosed to the subjects, and whether Institutional Review Board (IRB) approvals (or an equivalent approval/review based on the requirements of your country or institution) were obtained?
    \item[] Answer: \answerNA{} 
    \item[] Justification: No crowdsourcing or human subjects were involved in the experiments conducted for this paper.

\item {\bf Declaration of LLM usage}
    \item[] Question: Does the paper describe the usage of LLMs if it is an important, original, or non-standard component of the core methods in this research? Note that if the LLM is used only for writing, editing, or formatting purposes and does not impact the core methodology, scientific rigorousness, or originality of the research, declaration is not required.
    \item[] Answer: \answerNA{} 
    \item[] Justification: LLMs are not a core component of this paper in terms of methods.

\end{enumerate}

\newpage
\appendix

\newpage

\section{Supplementary Related Work}\label{app:supplementary_related_work}

Effective propagation and preservation of information on graphs remains a central challenge in deep learning on graphs, especially when long-range communication between nodes becomes fundamental for the downstream task \cite{shi2023exposition}. GNNs usually rely on local neighborhood aggregation, which limits their capacity to capture interactions between distant nodes \citep{alon2021on, di2023over} due to challenges such as over-smoothing \citep{cai2020note, oono2020graph, rusch2023survey} and over-squashing \citep{alon2021on, topping2022understanding, di2023over}, which are linked to the problem of vanishing gradients \citep{arroyo2025vanishing}.
Several techniques have been proposed to address this issue. 

Graph rewiring techniques \citep{DIGL,alon2021on,attali2024rewiringtechniquesmitigateoversquashing} modify the original topology, usually as a pre-processing step, with the aim of directly connect distant nodes and facilitate information flow. Rewiring methods can be broadly classified based on the type of information they leverage, including curvature metrics \citep{alon2021on,borf,pmlr-v231-fesser24a}, effective resistance \cite{black2023understanding}, random walks \cite{barbero2023locality}, spectral gap \cite{karhadkar2022fosr}, and node features \citep{gutteridge2023drew}.

Similarly, Graph Transformers \cite{shehzad2024graphtransformerssurvey,san,graphormer,rampavsek2022recipe} enable direct message passing between any pair of nodes via attention mechanisms, employing an attention mechanism on the entire graph. These models often incorporate positional encodings, such as Laplacian eigenvectors \citep{dwivedi2020generalization} or random-walk structural encodings (RWSE) \citep{dwivedi2022graph}, to encode graph structure. To reduce the quadratic complexity of the full attention mechanism, recent methods introduce methods such as sparse attention mechanisms \citep{Zaheer2020, Choromanski2020}, Exphormer \citep{shirzad2023exphormer}, and linear graph transformers \citep{deng2024polynormer}.

Despite the success of graph rewiring methods and Graph Transformers, these approaches often introduce additional computational complexity due to the use of denser graph shift operators. An alternative strategy to enhance long-range propagation focuses on constraining the dynamics of the GNN to remain stable and non-dissipative, while maintaining the computational complexity of classical MPNNs. In this paradigm, the GNN is interpreted as a discretization of a differential equation, leveraging dynamical systems theory to maintain a constant rate of information flow between nodes. This behavior has been achieved either through antisymmetric weight parameterization \cite{gravina_adgn, gravina_swan, gravina_randomized_adgn, gravina_ctan} or by exploiting port-Hamiltonian dynamics \cite{gravina_phdgn}.

Recent works further explore long-range interactions in GNNs. In \cite{bamberger2025measuring}, Bamberger et al. formalize and measure interaction ranges in graph operators, while in \cite{liang2025towards} Liang et al. introduce a large-scale dataset and metric to quantify long-range dependencies. These studies complement existing approaches by providing tools to better evaluate and understand information propagation across distant nodes.

Other approaches include filtering messages in the information flow \cite{finkelshtein2024cooperative,amp}, using a graph adaptive method based on a learnable ARMA framework \cite{grama}, state space models \cite{behrouz2024graph, mpssm}, or fractional power of the graph shift operator \cite{maskey2024fractional}.

\section{Sensitivity Results}\label{app:proofs}
In this section, we provide the proofs for the statements in \Cref{sec:signal-propagation}.

\subsection{Proof of \Cref{lem:linear_jacobian}}

\begin{proof}
Applying vectorization to $f(\bfX)$ and recalling that $\operatorname{vec}(\mathbf{A} \mathbf{X} \mathbf{B}) = (\mathbf{B}^\top \otimes \mathbf{A}) \operatorname{vec}(\mathbf{X})$, we obtain
\[ \operatorname{vec}(f(\mathbf{X})) = \sum_{k=1}^{K} (\mathbf{\Theta}_k^\top \otimes \mathbf{L}^k) \operatorname{vec}(\mathbf{X}).\] 
Taking derivatives:
\[
\mathbf{J}_{\boldsymbol{f}} = \frac{\partial \boldsymbol{f}}{\partial \operatorname{vec}(\mathbf{X})} =  \sum_{k=1}^{K} (\mathbf{\Theta}_k^\top \otimes T_k(\mathbf{L})).
\]
\end{proof}

\subsection{Proof of \Cref{lem:nonlinear_jacobian}}

\begin{proof}
Define the \(k\)-hop Jacobian block \(\mathbf{J}_k = \mathbf{\Theta}_k \otimes \mathbf{L}^{k}\)
and the full Jacobian \(\mathbf{J} = \sum_{k=1}^K \mathbf{J}_k\). Because
\(\mathbf{\Theta}_k \otimes \mathbf{L}^{k}\) has singular values
\(s_j(\mathbf{\Theta}_k)\,\lvert\lambda_i\rvert^{k}\),
the squared singular values of $\mathbf{J}_k$ are given by
\(\gamma^{(k)}_{i,j} = \lambda_i^{2k}\,\mu_{j,k}\). 

For a \emph{square} Gaussian matrix, the empirical spectrum of
\(\Theta_k\Theta_k^{\!\top}\) converges to the
Marchenko–Pastur (MP) law.
Scaling by \(\sigma^{2}\lambda_i^{2k}\) yields
\begin{align}\label{eq:moments_per_k}
  m^{(k)}_1 &= \sigma^{2}\lambda_i^{2k} \\
  m^{(k)}_2 &= 2\big(\sigma^{2}\lambda_i^{2k}\big)^2.
\end{align}
Independence, together with the rotational symmetry of each
\(\mathbf{\Theta}_k\) implies the blocks \(\mathbf{J}_k\) are
\emph{asymptotically free}.
Freeness gives additive \(R\)-transforms,
hence additive \emph{free} cumulants
\(\kappa_r\bigl(\sum_k \mathbf{J}_k\bigr) = \sum_k \kappa_r(\mathbf{J}_k)\).
For \(r=1,2\) the classical moments coincide with the free cumulants,
so the ordinary moments of \(\gamma_{i,j}\) also add:
\begin{equation}\label{eq:add_moments}
  m_r = \sum_{k=1}^{K} m^{(k)}_r,
  \qquad r\in\{1,2\}.
\end{equation}
  Insert \Cref{eq:moments_per_k} into \Cref{eq:add_moments}:
\begin{align}
    m_1 &= \sigma^{2}\sum_{k=1}^{K}\lambda_i^{2k} \\
    m_2 &= 2\sigma^{4}\sum_{k=1}^{K}\lambda_i^{4k}. 
\end{align}
 
Finally,
\begin{align*}
      Var[\gamma_{i,j}] &= m_2 - m_1^{2} \\
  &= 2\sigma^{4}\!\,\sum_{k}\lambda_i^{4k}
    -\sigma^{4}\Bigl(\sum_{k}\lambda_i^{2k}\Bigr)^{2} \\
  &= \sigma^{4}\bigl(\sum_{k}\lambda_i^{2k}\bigr)^{2}.
\end{align*}
\end{proof}

\subsection{Proof of \Cref{thm:chebnet_sensitivity}}


\begin{proof}
Recall that the forward pass for one  ChebNet layer (omitting the activation function) is:
\begin{equation}
    \mathbf{X}^{(l+1)} =  \sum_{k=0}^{K} T_k(\mathbf{L}) \mathbf{X}^{(l)} \mathbf{W}_k^{(l)}.
\end{equation}
Applying vectorization, we obtain:
\begin{equation}
    \text{vec}(\mathbf{X}^{(l+1)}) =  \sum_{k=0}^{K}\left((\mathbf{W}_k^{(l)})^\top \otimes T_k(\mathbf{L})\right)\text{vec}(\mathbf{X}^{(l)}).
    \label{eq:vec_eulerChebNet}
\end{equation}

By unrolling \Cref{eq:vec_eulerChebNet} and taking the derivative with respect to $\text{vec}(\mathbf{X}^{(0)})$, we obtain  the sensitivity after $l$ layers:
\begin{equation}
    \frac{\partial \text{vec}(\mathbf{X}^{(l)})}{\partial \text{vec}(\mathbf{X}^{(0)})} = \prod_{l=0}^{l-1}\left( \sum_{k=0}^{K}\left((\mathbf{W}_k^{(l)})^\top \otimes T_k(\mathbf{L})\right)\right).
\end{equation}

Focusing on a single feature channel (or summing across channels), we obtain the sensitivity:
\begin{equation}
    \frac{\partial \mathbf{X}^{(l)}}{\partial \mathbf{X}^{(0)}} = \prod_{l=0}^{l-1}\left( \sum_{k=0}^{K}T_k(\mathbf{L})\mathbf{W}_k^{(l)}\right).
\end{equation}

Then, the sensitivity of node $v$ with respect to node $u$ after $l$ layers is given by:
\begin{equation}
    \frac{\partial\mathbf{x}_v^{(l)}}{\partial\mathbf{x}_u^{(0)}} = \left(\prod_{l=0}^{l-1} \left(\sum_{k=0}^{K} T_k(\mathbf{L})\mathbf{W}_k^{(l)}\right)\right)_{v,u}.
\end{equation}
\end{proof}

\subsection{Proof of \Cref{thm:chebnetODE_eigenvalues}}
\begin{proof}
Given that the graph-wise Jacobian of \Cref{eq:chebnetODE} is \Cref{eq:spectral_gnn_jacobian}, we note that each term in the Jacobian, of the form $\bfW_k^\top \otimes T_k(\bfL)$, is an antisymmetric matrix. This follows from the fact that $\bfW_k$ is antisymmetric by construction, $T_k(\bfL)$ preserves the symmetry of the normalized Laplacian, and the Kronecker product of an antisymmetric matrix with a symmetric matrix is itself antisymmetric. Finally, since the sum of antisymmetric matrices remains antisymmetric, it follows that the graph-wise Jacobian of \Cref{eq:chebnetODE} has purely imaginary eigenvalues.
\end{proof}

\subsection{Sensitivity upperbound of standard MPNNs}\label{app:sensitivity_mpnns}
In the following theorem we report the sensitivity upperbound computed for standard MPNNs in \cite{di2023over}.

\begin{theorem}[Sensitivity uppperbound of standard MPNNs,  taken from \cite{di2023over}]\label{theo:sensitivity_digiovanni}
    Consider a standard MPNN with $l$ layers, where $c_\sigma$ is the Lipschitz constant of the activation $\sigma$, $w$ is the maximal entry-value over all weight matrices, and $d$ is the embedding dimension. For $u,v\in V$ we have
    \begin{equation}\label{eq:mpnn_over-squashing}
    \left\|\frac{\partial\mathbf{h}_v^{(l)}}{\partial\mathbf{h}_u^{(0)}}\right\| \leq \underbrace{(c_\sigma w d)^l}_{model}\underbrace{(\mathbf{O}^l)_{vu}}_{topology},
\end{equation}    
with $\mathbf{O}=c_r\mathbf{I}+c_a\mathbf{A}\in\mathbb{R}^{n\times n}$ is the message passing matrix adopted by the MPNN, with $c_r$ and $c_a$ are the contributions of the self-connection and aggregation term.
\end{theorem}

\subsection{Proof of \Cref{thm:Non-exponential_Information_Growth}}


\begin{proof}
First, recall the Jacobian explicitly:
\begin{equation}
    \bfJ^{(l)} = \mathbf{I} + \epsilon \sum_{k=0}^{K}\left((\mathbf{W}_k^{(l)})^\top \otimes T_k(\mathbf{L})\right).
\end{equation}
Define:
\begin{equation}
    A^{(l)} = \sum_{k=0}^{K}\left((\mathbf{W}_k^{(l)})^\top \otimes T_k(\mathbf{L})\right).
\end{equation}

\textbf{Antisymmetric Case:} When $(\mathbf{W}_k^{(l)})^\top = -\mathbf{W}_k^{(l)}$, the matrix $A^{(l)}$ is antisymmetric, because it is a Kronecker product of antisymmetric and symmetric matrices. Its eigenvalues are purely imaginary (or symmetric about zero), meaning their real parts vanish. Thus, for the spectral radius, we have:
\begin{align}
\|\bfJ^{(l)}\|_2^2 &= \rho\left((\bfJ^{(l)})^\top \bfJ^{(l)}\right)\\
&= \rho\left(\mathbf{I} + \epsilon(A^{(l)}+(A^{(l)})^\top) + \epsilon^2 (A^{(l)})^\top A^{(l)}\right).
\end{align}
Due to antisymmetry:
\begin{equation}
    (A^{(l)})^\top + A^{(l)} = 0.
\end{equation}
Hence:
\begin{equation}
    \|\bfJ^{(l)}\|_2^2 = \rho\left(\mathbf{I} + \epsilon^2 (A^{(l)})^\top A^{(l)}\right).
\end{equation}
The matrix $(A^{(l)})^\top A^{(l)}$ is symmetric and positive semi-definite, having nonnegative real eigenvalues. Expanding around the identity gives:
\begin{equation}
    \|\bfJ^{(l)}\|_2 = \sqrt{1 + \epsilon^2 \lambda_{\max}\left((A^{(l)})^\top A^{(l)}\right)} = 1 + O(\epsilon^2).
\end{equation}
Thus, no exponential growth or decay occurs.

\textbf{General Case of Stable-ChebNet (Without Antisymmetry):} For arbitrary matrices, the linear term $(A^{(l)}+(A^{(l)})^\top)$ typically does not vanish, introducing nonzero real eigenvalues. This causes exponential growth or decay in the Jacobian norm across layers:
\begin{equation}
    \|\bfJ^{(l)}\|_2 \approx 1 \pm C\epsilon,
\end{equation}
for some constant $C = \max_{i=0, \cdots, l} C^{(i)}>0$. Iterating over layers results in exponential instability:
\begin{equation}
    \|\bfJ^{(l)} \bfJ^{(l-1)} \dots \bfJ^{(0)}\|_2 \approx (1 \pm C\epsilon)^l,
\end{equation}
which grows or decays exponentially as the depth $l$ increases.

Thus, antisymmetric weights provide explicit protection against exponential growth or decay, while general weights typically do not.

\textbf{Standard ChebNet Case:} Iterating over layers, this results in exponential instability:
\begin{equation}
    \|\bfJ^{(l)} \bfJ^{(l-1)} \dots \bfJ^{(0)}\|_2 \approx (\pm C)^l,
\end{equation}
which grows or decays exponentially as the depth $l$ increases.

\end{proof}

\section{Dataset and Baseline Description} \label{app:datasets_and_baselines}
\subsection{Graph Property Prediction task description} \label{app:graphprop_description}

For the graph property prediction dataset, we make use of three separate tasks:

\textbf{Diameter (Graph-Level).} The diameter is defined as the length of the longest shortest path between any two nodes in the graph. It requires aggregating information from distant regions of the graph.

\textbf{SSSP (Node-Level).} Single-Source Shortest Path requires predicting each node’s distance to a designated source node. Solving this task with GNNs places a strong emphasis on propagating information from nodes that may lie many hops away from the source.

\textbf{Eccentricity (Node-Level).} For each node $u$, the eccentricity is the length of the maximum shortest path between $u$ and any other node. As with diameter and SSSP, accurate eccentricity estimation relies heavily on capturing long-distance relationships.

In total, 
the task contains 5,120 graphs for training, 640 for validation, and 1,280 for testing. The train/validation/test splits follow the same seed and 
setup in \cite{gravina_adgn}. We train our models by optimizing mean squared error (MSE) for each task, performing a grid search over hyperparameters (e.g., learning rate, weight decay, number of layers). Each experiment is repeated across four random initializations, and we report the average performance.

Because shortest-path-related tasks inherently rely on propagating signals over large portions of a graph, they serve as a natural stress test for the capacity of GNNs to perform long-range propagation. 

\subsection{Barbell graph task description} \label{app:barbel_description}
\begin{figure}[H]
    \centering
    \includegraphics[width=0.65\textwidth]{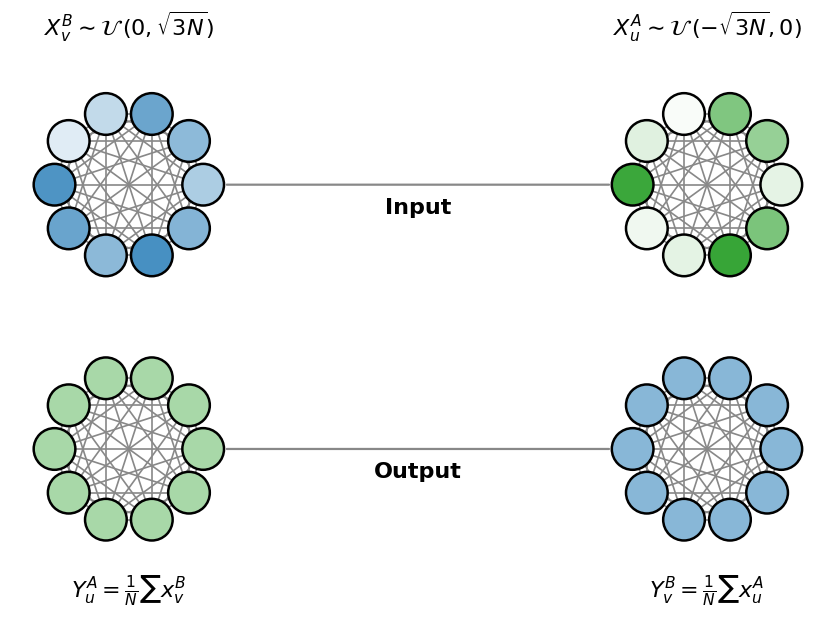}
    \caption{Illustration of input and output node features in a barbell graph setup. Adapted from Bamberger et.al \cite{bamberger2024bundle}.}
    \label{fig:barbell}
\end{figure}

A Barbell graph $B_{n,k}$ is formed by connecting two complete graphs $K_n$ (the ``bells'') with a simple path of length $k$ (the ``bridge''). Therefore, every node inside a bell has high intra-cluster connectivity ($n - 1$ neighbors), whereas the bridge nodes have degree 2 and constitute the only communication route between the two bells. In this task, the target for nodes is to output the average input feature
over nodes of the opposite bell and vice-versa, as illustrated in \Cref{fig:barbell}. \textbf{Mean squared error (MSE)} is used for node-level regression as a proxy for how severely the GNN is either \textit{oversquashing} (failing to pass information across the narrow ``bridge'') or \textit{oversmoothing} (collapsing all node embeddings to be nearly identical). Numerical MSE outcomes are associated with each pathology. 

\textbf{An MSE of around 1 corresponds to oversquashing:}
The model is so bottlenecked by the bridge that it effectively ignores or fails to incorporate information from the other ``bell.'' In other words, each side of the barbell can predict its own node labels, but the information from the opposite side never gets through, leading to a characteristic level of error ($\approx 1$ in their chosen label distribution).

\textbf{An MSE of around 30 corresponds to oversmoothing:} The model passes messages so many times (or in such a way) that it ``collapses'' all node embeddings toward the same prediction, ignoring local distinctions within each side. Because the barbell's node labels are diverse (randomly assigned), forcing every node toward the same value yields a much larger overall MSE ($\approx 30$ in their setup).


On the other hand, an MSE of around 0.5 as shown in the results simply means that the model is doing better than the severe oversquashing case (it is not completely failing to pass information across the barbell). In other words, some amount of meaningful communication is happening between the two ``bells,'' and the node representations are not entirely collapsed.

\subsection{Employed baselines}\label{app:baselines_details}
In our experiments, the performance of our method is compared with various state-of-the-art GNN baselines from the literature. Specifically, we consider:
\begin{itemize}
\item Classical GNN methods, \ie GCN~\citep{kipf2016semi}, GraphSAGE~\citep{hamilton2017inductive}, GAT~\citep{velivckovic2018graph}, GIN~\citep{xu2019powerful}, and GCNII~\citep{gcnii}, ChebNet~\citep{defferrard2016convolutional}, ChebNetII~\citep{he2022convolutional}, CIN++~\citep{giusti2023cinenhancingtopologicalmessage}, PathNN~\citep{pathnn}, S2GCN~\citep{geisler2024spatio}, SGC~\citep{SGC}, SIGN~\citep{frasca2020signscalableinceptiongraph}, GRAMA \cite{grama},
GatedGCN~\citep{gatedgcn},
CoGNN~\citep{finkelshtein2024cooperative}, 
H2GCN \citep{h2gcn}, 
CPGNN \citep{CPGNN}, 
FAGCN \citep{FAGCN}, 
GPR-GNN \citep{GPR_GNN},
FSGNN \citep{FSGNN}, 
GloGNN \cite{GloGNN}, 
GBK-GNN \citep{GBK_GNN}, 
JacobiConv \citep{jacobiconv};
\item Differential-equation inspired GNNs (DE-GNNs), \ie DGC~\citep{DGC}, GRAND~\citep{chamberlain2021grand}, GraphCON~\citep{rusch2022graph}, A-DGN~\citep{gravina_adgn}, and SWAN~\citep{gravina_swan}
PH-DGN \cite{gravina_phdgn};
\item Graph Transformers, \ie SAN~\citep{san}, GraphGPS~\citep{rampavsek2022recipe}, GOAT~\citep{goat}, Exphormer~\citep{shirzad2023exphormer}, {GRIT}\citep{grit}, {GraphViT} \citep{he2023generalization}, G.MLPMixer~\citep{he2023generalization}
SPEXPHORMER~\citep{shirzad2024even}, TIGT~\citep{choi2025topologyinformedgraphtransformer}, SGFormer~\citep{wu2023sgformer}, NodeFormer~\citep{wu2022nodeformer}, Specformer~\citep{bo2023specformer}, GCN-nsampler and GAT-nsampler~\citep{zeng2019graphsaint},

GT~\citep{ijcai2021p0214},
NAGphormer \cite{nagphormer},
Polynormer~\cite{deng2024polynormer};
\item Rewiring-based methods, \ie LASER~\citep{barbero2023locality}, and DRew~\citep{gutteridge2023drew};
\item SSM-based GNN, \ie Graph-Mamba~\citep{wang2024graph}, GMN~\citep{behrouz2024graph},
MP-SSM \cite{mpssm}
\end{itemize}

\section{Hyper-parameter grids}

\Cref{tab:hyperparam_grid_peptides,tab:hyperparam_grid_graphprop,tab:hyperparam_grid_ogb} summarize our hyperparameter exploration: Table \ref{tab:hyperparam_grid_peptides} listing the sweep ranges for Stable-ChebNet on Peptides-func, graph-property benchmarks (Diameter, SSSP, Eccentricity), and ogbn-arxiv along with ogbn-proteins experiments, respectively.
\label{append:hyperparams}

\begin{table}[H]
\centering
\caption{Hyper-parameter grid for Stable-ChebNet ablation on \textit{Peptides-func}.}
\label{tab:hyperparam_grid_peptides}
\begin{tabular}{lcc}
\toprule
\textbf{Hyper-parameter} & \textbf{Reference} & \textbf{Sweep values used in ablation} \\
\midrule
Hidden dim $d$ & \textbf{140} & 100, 120, 140, 145, 160 \\
Polynomial order $K$ & \textbf{10} & 6,8,10 \\
Num of layers & \textbf{4} & 3,4,5 \\
MLP layers & \textbf{2} & 1,2,3 \\
Step size $\varepsilon$ & \textbf{0.5} & $[0.1, 1.0]$ \\
Dissipative force $\gamma$ & \textbf{0.05} & 0.001, 0.01, 0.05, 0.1 \\
Batch size & \textbf{64} & 32, 64, 128 \\
Learning rate & $\mathbf{0.001}$ & $0.0001,0.001,0.01$ \\
Optimizer & AdamW & AdamW \\
Pos-enc type & None & None, Laplacian, RW \\
Pos-enc dim & \textbf{16} & 8, 16, 32 \\
\bottomrule
\end{tabular}
\end{table}


\begin{table}[h]
\centering
\small
\caption{Hyper‑parameter grid and best settings for Stable‑ChebNet on three synthetic graph‑property benchmarks.}
\label{tab:hyperparam_grid_graphprop}
\begin{tabular}{lcccc}
\toprule
\textbf{Hyper‑parameter} & \textbf{Values in grid} & \textbf{Diam} & \textbf{SSSP} & \textbf{Ecc} \\
\midrule
Hidden dimension $d$            & 20, 30, 50                       & 50  & 30  & 30  \\
Number of layers                & 1, 2, 3, 5, 10, 20               & 20   & 5   & 5   \\
Polynomial order $K$            & 3, 5, 10                    & 4  & 10  & 10  \\
Step size $\varepsilon$         & 0.01, 0.10, 0.20, 0.30      & 0.40& 0.30& 0.30\\
Dissipative force $\gamma$      & 0, 0.01, 0.50, 1         & 0.01& 0.00& 0.00\\
Activation function             & \texttt{tanh}, \texttt{relu}& \texttt{relu}& \texttt{relu}& \texttt{relu}\\
Learning rate                   & 0.001,0.003                       & 0.003 & 0.003 & 0.003 \\
Weight decay                    & $1\times10^{-6}$            & $1\times10^{-6}$ & $1\times10^{-6}$ & $1\times10^{-6}$\\
\bottomrule
\end{tabular}
\end{table}

\begin{table}[h]
\centering
\caption{Hyper-parameter sweep ranges for Stable-ChebNet on \texttt{ogbn-arxiv} and \texttt{ogbn-proteins}.}
\label{tab:hyperparam_grid_ogb}
\begin{tabular}{lcc}
\toprule
\textbf{Hyper-parameter} & \textbf{Sweep (arxiv)} & \textbf{Sweep (proteins)} \\
\midrule
Hidden dim $d$ & 128, 256, 512 & 256, 512, 1024 \\
Polynomial order $K$ & 4, 5, 6, 10 & 5, 10, 15 \\
Num of layers & 2, 3, 4, 5 & 3, 5, 7 \\
MLP layers & 1, 2, 3 & 1, 2, 3 \\
Step size $\epsilon$ & [0.1, 1.0] & [0.1, 1.0] \\
Dissipative force $\gamma$ & 0.01, 0.05, 0.1 & 0.01, 0.05, 0.1 \\
Batch size & 256, 512, 1024 & 512, 1024, 2048 \\
Learning rate & 0.001, 0.01, 0.05 & 0.0005, 0.001, 0.005 \\
Optimizer & Adam & Adam \\
Pos-enc type & None, Laplacian, RW & None, Laplacian, RW \\
Pos-enc dim & 8, 16, 32 & 16, 32, 64 \\
\bottomrule
\end{tabular}
\end{table}

\section{Eigenvalues distribution comparison}\label{app:eigenvalue_distribution}

\begin{figure}[htbp]
  \centering

  \begin{subfigure}[b]{\textwidth}
    \centering
    \includegraphics[width=0.9\textwidth]{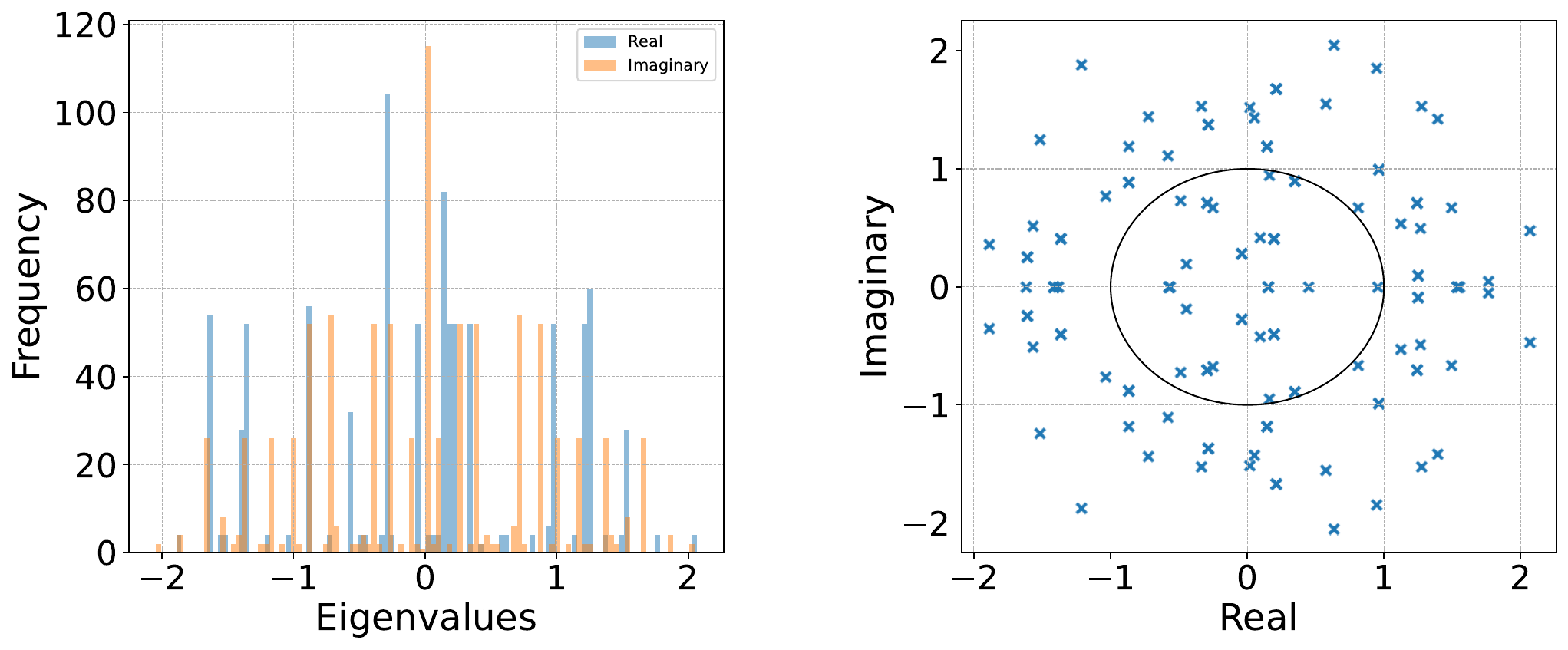}
    \caption{}
    \label{fig:chebneteig}
  \end{subfigure}

  \vspace{0.5cm} 

  \begin{subfigure}[b]{\textwidth}
    \centering
    \includegraphics[width=0.9\textwidth]{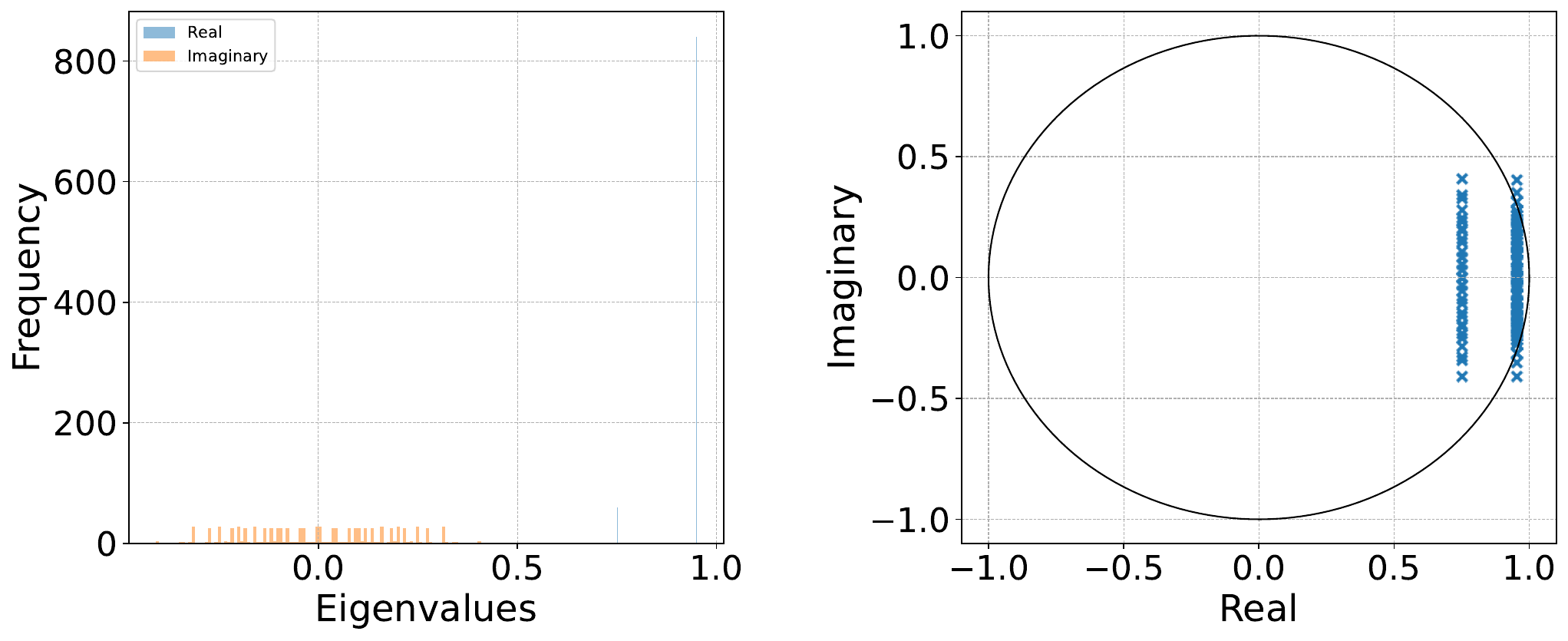}
    \caption{}
    \label{fig:stablechebneteig}
  \end{subfigure}

  \caption{Comparison of eigenvalue distributions: (a) ChebNet and (b) StableChebNet.}
  \label{fig:eigenvaluesplot}
\end{figure}

Figures \ref{fig:chebneteig} and \ref{fig:stablechebneteig} provide an empirical comparison of the Jacobian eigenvalue spectra for a classical ChebNet versus our Stable-ChebNet layer at $K=5$. The top two panels illustrate ChebNet’s spectrum: the left histogram shows that both the real and imaginary parts of its eigenvalues span broadly from roughly $-2$ to $+2$, with pronounced peaks near the extremes signaling a large spectral radius and a susceptibility to unstable, oscillatory dynamics. The accompanying scatter plot maps these eigenvalues in the complex plane, revealing many points lying well outside the unit circle, which corroborates the theoretical prediction that high-order Chebyshev filters can push the system toward chaotic regimes.

In contrast, the bottom row portrays the spectrum of Stable-ChebNet. Its histogram (bottom left) is tightly concentrated within approximately $-0.4$ to $+0.4$ on both axes, indicating that eigenvalues remain far inside the unit circle. The complex-plane scatter (bottom right) further demonstrates that all eigenvalues lie symmetrically about the imaginary axis and are bounded in magnitude, consistent with the antisymmetric weight parameterization, guaranteeing purely imaginary Jacobian eigenvalues and therefore more stable dynamics.

\section{Additional Experiments on Heterophilic Benchmarks}\label{app:hetero}
To further evaluate the performance of our Stable-ChebNet, we assess its the effectiveness in capturing complex relational information in heterophilic settings, where nodes belonging to same class are often connected through longer and sparser paths, we consider the five node classification tasks introduced in \cite{platonov2023a}. Specifically, we consider the  ``Roman-empire'', ``Amazon-ratings'' and ``Minesweeper'', ``Tolokers'' datasets. We adhere to the same data and experimental setting presented in \cite{platonov2023a}.

\begin{table}[h!]
\setlength{\tabcolsep}{3pt}
\centering
\caption{Mean test set score and std averaged over 4 random weight initializations on heterophilic datasets. The higher, the better.}
\label{tab:results_hetero_complete}
\scriptsize
\begin{tabular}{lcccc}
\hline\toprule
\multirow{2}{*}{\textbf{Model}} & \textbf{Roman-empire} & \textbf{Amazon-ratings} & \textbf{Minesweeper} & \textbf{Tolokers}\\
& \scriptsize{Acc $\uparrow$} & \scriptsize{Acc $\uparrow$} & \scriptsize{AUC $\uparrow$} & \scriptsize{AUC $\uparrow$}\\
\midrule
\multicolumn{4}{l}{\textbf{\cite{luan2024heterophilicgraphlearninghandbook}}} \\
$\,$ MLP-2 & 66.04$_{\pm0.71}$ & 49.55$_{\pm0.81}$ & 50.92$_{\pm1.25}$  & 74.58$_{\pm0.75}$\\
$\,$ SGC-1 & 44.60$_{\pm0.52}$ & 40.69$_{\pm0.42}$ & 82.04$_{\pm0.77}$  & 73.80$_{\pm1.35}$\\
$\,$ MLP-1 & 64.12$_{\pm0.61}$ & 38.60$_{\pm0.41}$ & 50.59$_{\pm0.83}$  & 71.89$_{\pm0.82}$\\
\midrule
\textbf{MPNNs} \\
$\,$ GAT         & 80.87$_{\pm0.30}$ & 49.09$_{\pm0.63}$ & 92.01$_{\pm0.68}$ & 83.70$_{\pm0.47}$\\
$\,$ GAT (LapPE) & 84.80$_{\pm0.46}$ & 44.90$_{\pm0.73}$ & 93.50$_{\pm0.54}$ & 84.99$_{\pm0.54}$\\
$\,$ GAT (RWSE)  & 86.62$_{\pm0.53}$ & 48.58$_{\pm0.41}$ & 92.53$_{\pm0.65}$ & 85.02$_{\pm0.67}$\\
$\,$ Gated-GCN   & 74.46$_{\pm0.54}$ & 43.00$_{\pm0.32}$ & 87.54$_{\pm1.22}$ & 77.31$_{\pm1.14}$\\
$\,$ GCN         & 73.69$_{\pm0.74}$ & 48.70$_{\pm0.63}$ & 89.75$_{\pm0.52}$ & 83.64$_{\pm0.67}$\\
$\,$ GCN (LapPE) & 83.37$_{\pm0.55}$ & 44.35$_{\pm0.36}$ & 94.26$_{\pm0.49}$ & 84.95$_{\pm0.78}$\\
$\,$ GCN (RWSE)  & 84.84$_{\pm0.55}$ & 46.40$_{\pm0.55}$ & 93.84$_{\pm0.48}$ & 85.11$_{\pm0.77}$\\
$\,$ CO-GNN($\Sigma$, $\Sigma$) & 91.57$_{\pm0.32}$ & 51.28$_{\pm0.56}$ & 95.09$_{\pm1.18}$ & 83.36$_{\pm0.89}$\\
$\,$ CO-GNN($\mu$, $\mu$) & 91.37$_{\pm0.35}$ & 54.17$_{\pm0.37}$ & 97.31$_{\pm0.41}$ & 84.45$_{\pm1.17}$\\
$\,$ SAGE        & 85.74$_{\pm0.67}$ & 53.63$_{\pm0.39}$ & 93.51$_{\pm0.57}$ & 82.43$_{\pm0.44}$\\
\midrule
\textbf{Graph Transformers} \\
$\,$ Exphormer   & 89.03$_{\pm0.37}$ & 53.51$_{\pm0.46}$ & 90.74$_{\pm0.53}$ & 83.77$_{\pm0.78}$\\
$\,$ NAGphormer  & 74.34$_{\pm0.77}$ & 51.26$_{\pm0.72}$ & 84.19$_{\pm0.66}$ & 78.32$_{\pm0.95}$\\
$\,$ GOAT        & 71.59$_{\pm1.25}$ & 44.61$_{\pm0.50}$ & 81.09$_{\pm1.02}$ & 83.11$_{\pm1.04}$\\
$\,$ GPS         & 82.00$_{\pm0.61}$ & 53.10$_{\pm0.42}$ & 90.63$_{\pm0.67}$ & 83.71$_{\pm0.48}$\\
$\,$ GPS\textsubscript{GCN+Performer} (LapPE) & 83.96$_{\pm0.53}$ & 48.20$_{\pm0.67}$ & 93.85$_{\pm0.41}$ & 84.72$_{\pm0.77}$\\
$\,$ GPS\textsubscript{GCN+Performer} (RWSE) & 84.72$_{\pm0.65}$ & 48.08$_{\pm0.85}$ & 92.88$_{\pm0.50}$ & 84.81$_{\pm0.86}$\\
$\,$ GPS\textsubscript{GCN+Transformer} (LapPE) & OOM & OOM & 91.82$_{\pm0.41}$ & 83.51$_{\pm0.93}$\\
$\,$ GPS\textsubscript{GCN+Transformer} (RWSE)  & OOM & OOM & 91.17$_{\pm0.51}$ & 83.53$_{\pm1.06}$\\
$\,$ GT          & 86.51$_{\pm0.73}$ & 51.17$_{\pm0.66}$ & 91.85$_{\pm0.76}$ & 83.23$_{\pm0.64}$\\
$\,$ GT-sep      & 87.32$_{\pm0.39}$ & 52.18$_{\pm0.80}$ & 92.29$_{\pm0.47}$ & 82.52$_{\pm0.92}$\\
$\,$ Polynormer	 & 92.55$_{\pm0.30}$ & \textbf{54.81$_{\pm0.49}$} & 97.46$_{\pm0.36}$ & 85.91$_{\pm0.74}$\\
\midrule
\multicolumn{4}{l}{\textbf{Heterophily-Designated GNNs}} \\
$\,$ CPGNN       & 63.96$_{\pm0.62}$ & 39.79$_{\pm0.77}$ & 52.03$_{\pm5.46}$ & 73.36$_{\pm1.01}$\\
$\,$ FAGCN       & 65.22$_{\pm0.56}$ & 44.12$_{\pm0.30}$ & 88.17$_{\pm0.73}$ & 77.75$_{\pm1.05}$\\
$\,$ FSGNN       & 79.92$_{\pm0.56}$ & 52.74$_{\pm0.83}$ & 90.08$_{\pm0.70}$ & 82.76$_{\pm0.61}$\\
$\,$ GBK-GNN     & 74.57$_{\pm0.47}$ & 45.98$_{\pm0.71}$ & 90.85$_{\pm0.58}$ & 81.01$_{\pm0.67}$\\
$\,$ GloGNN      & 59.63$_{\pm0.69}$ & 36.89$_{\pm0.14}$ & 51.08$_{\pm1.23}$ & 73.39$_{\pm1.17}$\\
$\,$ GPR-GNN     & 64.85$_{\pm0.27}$ & 44.88$_{\pm0.34}$ & 86.24$_{\pm0.61}$ & 72.94$_{\pm0.97}$\\
$\,$ H2GCN       & 60.11$_{\pm0.52}$ & 36.47$_{\pm0.23}$ & 89.71$_{\pm0.31}$ & 73.35$_{\pm1.01}$\\
$\,$ JacobiConv  & 71.14$_{\pm0.42}$ & 43.55$_{\pm0.48}$ & 89.66$_{\pm0.40}$ & 68.66$_{\pm0.65}$\\
\midrule
\textbf{Graph SSMs} \\
$\,$ GMN          & 87.69$_{\pm0.50}$ & 54.07$_{\pm0.31}$ & 91.01$_{\pm0.23}$ & 84.52$_{\pm0.21}$\\
$\,$ GPS + Mamba  & 83.10$_{\pm0.28}$ & 45.13$_{\pm0.97}$ & 89.93$_{\pm0.54}$ & 83.70$_{\pm1.05}$\\
$\,$ {GRAMA}$_{\textsc{GCN}}$ & 88.61$_{\pm0.43}$ & 53.48$_{\pm0.62}$ & 95.27$_{\pm0.71}$ & \textbf{86.23$_{\pm1.10}$}\\
$\,$ {MP-SSM}     & 90.91$_{\pm0.48}$ & 53.65$_{\pm0.71}$ & 95.33$_{\pm0.72}$ & 85.26$_{\pm0.93}$\\
\midrule
\textbf{Ours} \\
$\,$ Stable-ChebNet & \textbf{92.03$_{\pm0.85}$} & 53.15$_{\pm0.21}$ & \textbf{95.71$_{\pm2.26}$} & 85.55$_{\pm3.35}$\\
\bottomrule\hline      
\end{tabular}
\end{table}

\end{document}